\documentclass[10pt,twocolumn,letterpaper]{article}
\usepackage{iccv}
\usepackage{times}
\usepackage{epsfig}
\usepackage{graphicx}
\usepackage{amsmath}
\usepackage{amssymb}

\usepackage{tabularx,booktabs,multirow}
\usepackage{mathtools}
\usepackage{ragged2e}
\graphicspath{{figures/}}
\usepackage[pagebackref=true,breaklinks=true,letterpaper=true,colorlinks,bookmarks=false]{hyperref}
\iccvfinalcopy 
\def\iccvPaperID{7925} 

\ificcvfinal\fi
\begin{document}
\title{Spectral Leakage and Rethinking the Kernel Size in CNNs}

\author{Nergis Tomen, $\,\,$ Jan C. van Gemert \\
Computer Vision Lab, Delft University of Technology\\
Delft, 2628CD, Netherlands \\
{\tt\small \{n.tomen,j.c.vangemert\}@tudelft.nl}
}

\maketitle
\ificcvfinal\thispagestyle{empty}\fi
\begin{abstract}
   Convolutional layers in CNNs implement linear filters which decompose the input into different frequency bands. However, most modern architectures neglect standard principles of filter design when optimizing their model choices regarding the size and shape of the convolutional kernel. In this work, we consider the well-known problem of spectral leakage caused by windowing artifacts in filtering operations in the context of CNNs. We show that the small size of CNN kernels make them susceptible to spectral leakage, which may induce performance-degrading artifacts. To address this issue, we propose the use of larger kernel sizes along with the Hamming window function to alleviate leakage in CNN architectures. We demonstrate improved classification accuracy on multiple benchmark datasets including Fashion-MNIST, CIFAR-10, CIFAR-100 and ImageNet with the simple use of a standard window function in convolutional layers. Finally, we show that CNNs employing the Hamming window display increased robustness against various adversarial attacks. Code is available\footnote{Code will be made available at github.}.
\end{abstract}
\vspace{-0.2cm}
\section{Introduction}
\begin{figure}[t]
    \hspace{-0.2cm}
    \footnotesize
    \begin{tabularx}{\linewidth}{cc} \toprule
        \hspace{0.3cm} Example Gabor filter &
        \hspace{-0.7cm} Learned CNN filter \\
        \cmidrule(lr){1-1} \cmidrule(lr{20pt}){2-2}
        \includegraphics[trim=130 10 700 30,clip, width=0.49\linewidth]{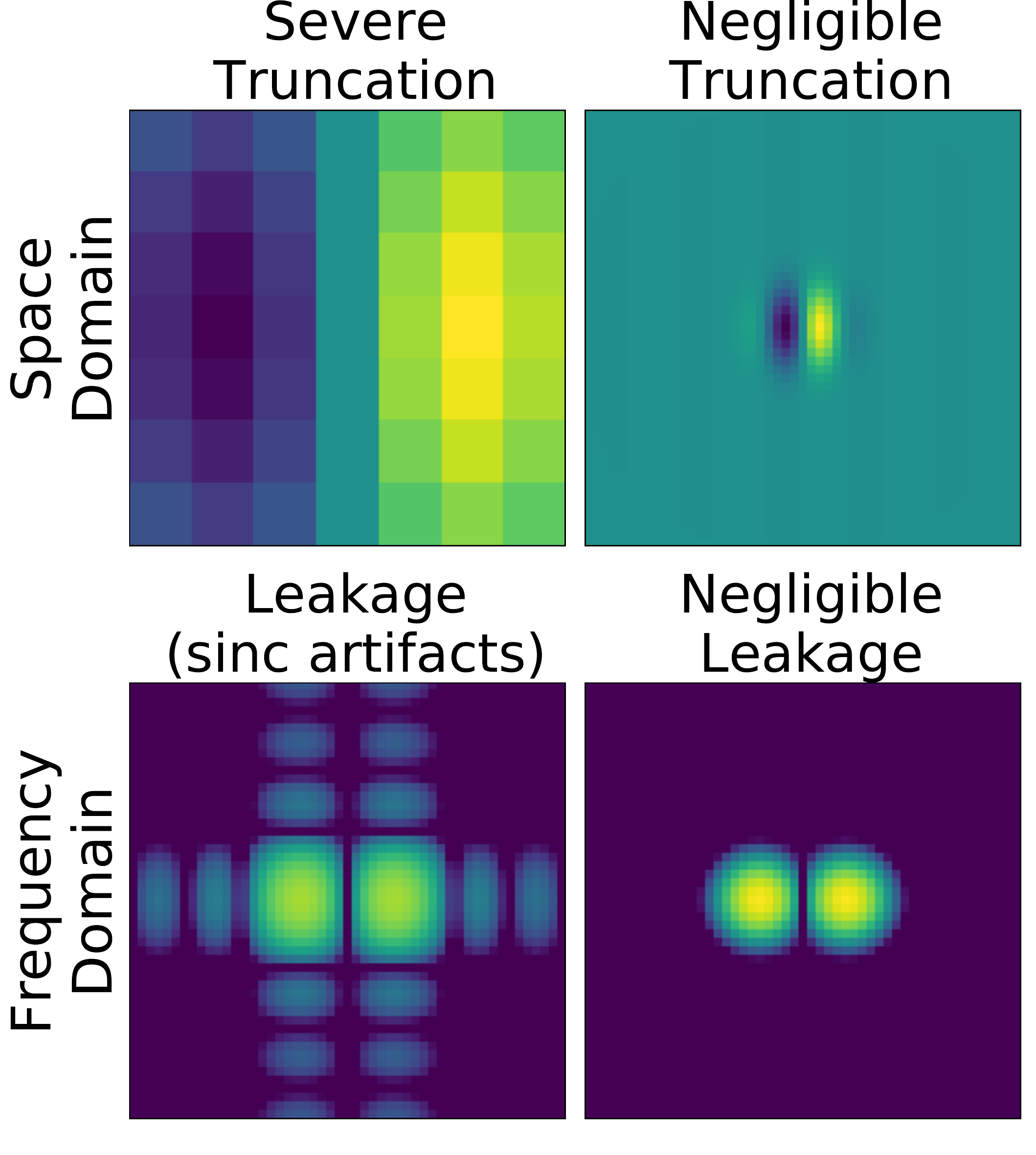} &
        \includegraphics[trim=740 10 90 30,clip, width=0.49\linewidth]{fig1.pdf} \\ \bottomrule
    \end{tabularx}
    \hspace{0.2cm}
    \caption{Windowing artifacts cause spectral leakage in the frequency domain, when a filter is not tapered off at the boundaries in the space domain.
    The top row shows example filters in space domain, while the bottom row are their corresponding frequency domains. Left: Example Gabor (bandpass) filter with severe truncation (kernel size $7 \times 7$) leads to spectral leakage in its frequency response due to sinc artifacts. The same filter with negligible truncation (kernel size $49 \times 49$) is a good quality bandpass filter. Right: A standard 7x7 CNN kernel trained on CIFAR-10 struggles to learn good quality bandpass filters, as the use of small kernel sizes typically lead to severe truncation. We propose using the standard Hamming window to taper off the kernels in space domain, which enables good quality bandpass frequency responses.}
    \label{fig:fig1}
\end{figure}


A fundamental component in deep image recognition networks is the ability to non-linearly stack learned, shareable, linear mappings. The canonical example is the linear convolution operator in CNNs~\cite{He2015,Krizhevsky2012,Simonyan2015}, while in recent visual Transformer models the query, key and values are token-shared linear mappings acting on pixel embeddings~\cite{cordonnier2019relationship,dosovitskiyICLR21VisionTransformers}. These linear mappings, which in the visual domain typically take the form of filters, allow image feature learning. Such learnable, hierarchical, shareable, feature detectors are fundamental to the great success of deep learning~\cite{Allen2020,Chen2020,Lee2009}, and a better understanding of these filters may broadly impact the whole field.


Image filters, such as local, oriented edge detectors, provide a highly reusable decomposition of the input image~\cite{Deza2020,Olshausen1996,Simoncelli2001} and are accurate models of early biological vision~\cite{Jones1987}. From a deep, hierarchical feature learning perspective, it is interesting to ask how specialized reusable filters should be, and explore their response selectivity. In particular, here we investigate the role of  frequency-selectivity of learned filters in deep networks. 


To investigate frequency-selectivity, we consider spectral leakage---which is a well-known~\cite{Hamming1998,Oppenheim1999,Prabhu2014} artifact in generic filtering operations---in the context of CNNs.
The building block of CNNs, the convolution operator, can be thought of as a linear filter, which, due to the small kernel sizes employed in modern CNNs, is susceptible to spectral leakage.
Although a well-studied concept in digital signal~\cite{Oppenheim1999} and image processing~\cite{Gonzalez2008,Jahne2005}, we observe that a broader understanding of spectral leakage in deep networks has largely been neglected.

Spectral leakage, in the broad sense of the term, is when an operation on a signal introduces unwanted frequency components to the result of that operation. In practice, the term leakage is typically used when a filter lets through frequency components of a signal outside of its intended passband due to windowing artifacts.
For linear filters implemented via discrete convolution or cross-correlation operators, a kernel with finite size can be interpreted as a truncated version of an infinite, ideal filter. The finite size of the discrete kernel, within which the filter assumes non-zero values, represents a multiplication of an infinite kernel and a rectangular function in space domain, which translates as a convolution with a sinc function in frequency domain~\cite{Prabhu2014}. When a two-dimensional bandpass filter, such as a Gabor, is severely truncated, the rectangular function introduces windowing artifacts to the frequency response in the form of `ripples' of the sinc function (Fig.~\ref{fig:fig1}, left).

Here we explore the effect of leakage artifacts in image classification performance in CNNs. We show that due to the typical choice of small kernel sizes, CNNs have little freedom to avoid rectangular truncation functions, which make them susceptible to spectral leakage (Fig.~\ref{fig:fig1}, right). We investigate the impact of leakage artifacts on benchmark classification tasks and demonstrate that the simple use of a standard window function which reduces leakage can improve classification accuracy. Furthermore, we show that windowed CNNs are more robust against certain types of adversarial examples.

Our contributions can be summarized as:
\vspace{-0.15cm}
\begin{itemize}
\item We investigate the impact of spectral leakage in CNNs. Although spectral leakage is a fundamental concept in classical signal processing, its impact on CNN performance has not been explicitly explored before.
\vspace{-0.15cm}
\item We employ principles of good filter design, which are largely ignored in CNN models, to propose the use of larger kernels with the standard Hamming window function, which is tapered off at the kernel boundaries to alleviate spectral leakage.
\vspace{-0.15cm}
\item We demonstrate improvements to classification accuracy in benchmark datasets including Fashion-MNIST, CIFAR-10, CIFAR-100 and ImageNet with the simple use of a standard window function.
\vspace{-0.15cm}
\item We show that windowed CNNs display increased robustness against certain adversarial attacks including DeepFool and spatial transformation attacks.
\end{itemize}

\section{Related work}
\subsection{Spectral leakage in signal processing}
A signal observed within a finite window with aperiodic boundary conditions may be seen as a longer signal, truncated by multiplication with a rectangular window. This truncation introduces sidelobes, or `ripples', to the spectral density of the signal and may, for example, decrease the signal-to-noise ratio in transmissions~\cite{Oppenheim1999}. Often referred to as spectral leakage, such windowing artifacts are closely related to the Gibbs phenomenon~\cite{Hamming1998} or ringing artifacts in digital image processing~\cite{Gonzalez2008}. To combat the undesirable effects of leakage, window functions are commonly employed in many applications including spectral analysis via short-time Fourier transforms~\cite{Prabhu2014}.

The use of 2-dimensional window functions is also commonplace for spectral decomposition in digital~\cite{Jahne2005} and biomedical~\cite{Semmlow2014} image processing.
Window functions are also an integral part of filter design, both in temporal~\cite{Prabhu2014} and image domains~\cite{Birchfield2016}. In this work, we consider the fundamental lack of window functions in CNN architectures and investigate whether leakage artifacts may lead to adverse effects. We propose the use of a Hamming window based on its reasonable attenuation of sidelobes while maintaining a relatively narrow main lobe.

\subsection{Signal processing benefits for CNNs}
Incorporating signal processing knowledge brought great benefits to deep learning. Marrying convolution to deep learning yields the CNN~\cite{MNIST}, the importance of which is difficult to overstate. The convolution theorem allows efficient CNN training~\cite{mathieuICLR14fastTrainFFTS}. The scattering transform~\cite{bruna2013scattering} and its variants~\cite{oyallon2015rotoTranslationScattering,oyallon2018scattering,Sifre2013} allow encoding domain knowledge. Structuring filters based on scale-space theory~\cite{jacobsenCVPR16structuredRF,Shelhamer2019} brings data efficiency. Anti-aliasing in CNNs~\cite{Zhang2019} increases robustness and accuracy.  Reducing border effects in CNNs~\cite{kayhan2020translation} improves translation equivariance and data efficiency. In this paper, we build on these successes and for the first time investigate spectral leakage in CNNs, where reducing spectral leakage increases classification accuracy and improves robustness.

\subsection{Kernel size and shape in CNNs}
In standard convolutional layers, using small kernel sizes reduces computational complexity and, empirically, improves accuracy, therefore increasingly small kernel sizes have been adopted over time in CNNs~\cite{He2015,Krizhevsky2012,Simonyan2015,Szegedy2015,Szegedy2016,Zeiler2014}.
One potential problem with larger kernel sizes may be over-parametrization, and our windowing method functions as a form of regularization which effectively constrains the parameter space. Unlike common regularization methods in deep learning, such as weight decay~\cite{Hanson1988}, dropout~\cite{Srivastava2014},
early stopping~\cite{Morgan1989} and data augmentation~\cite{Shorten2019}, our adopted Hamming window represents a spatially well-structured form of regularization which encourages a center-bias in the kernel shape. In fact, we show that our method is not a substitute for other types of regularization and synergizes well with weight decay and data augmentation.

Similarly, enforcing a center-bias in CNN filters is in line with the idea that images present hierarchically local problems~\cite{Deza2020,Olshausen1996,Simoncelli2001}, best tackled by local~\cite{Loog2017}, deep, hierarchical learning~\cite{Allen2020,Chen2020,Lee2009}. However, unlike previous work, we suggest that the use of small kernels may not be sufficient without explicit regularization of kernel boundaries.


Finally, other work has addressed structured kernel shapes in CNNs, using filter banks based on wavelets~\cite{bruna2013scattering}, Gabor filters~\cite{Luan2017}, Gaussian derivatives~\cite{jacobsenCVPR16structuredRF} and circular harmonics~\cite{Worrall2017}. Here, we focus explicitly on the truncation properties of the window function while learning the pixel weights in the conventional manner. This is similar to the approach of blurring the CNN filters~\cite{Shelhamer2019}, in our approach, however, we are  not blurring the filters, or the feature maps~\cite{Zhang2019}. Instead, we are performing a simple multiplication in space domain, unlike previous work.

\subsection{Adversarial attacks}
The features learned by deep models may not be robust which has implications for AI safety~\cite{Biggio2013,Goodfellow2014}. An important observation in CNNs is that adversarial images with perturbations tiny in magnitude or imperceptible by humans~\cite{Deepfool,Szegedy2013,Eykholt2018,Sharif2016}
can lead to misclassification with high confidence. Attacks can be generated based on information about the activations and gradients of the targeted network (white-box attacks)~\cite{Goodfellow2014}. However, many adversarial examples are highly transferable between models~\cite{Liu2017} and without access to model parameters (black-box attacks), substitute models~\cite{Papernot2017} or even simple spatial transformations~\cite{Engstrom2019} can be employed to generate adversarial images.

Adversarial examples can be traced back to brittle features which are inherent in the data distribution and are highly class-specific~\cite{Ilyas2019}. This means, that the reliance of classification models on very small magnitude features may boost performance, leading to a robustness-accuracy trade-off~\cite{Tsipras2019}.
Similarly, leakage artifacts are typically small in magnitude, but may be present in every feature map in a standard CNN with non-tapered filters. Here, we investigate whether simply using a tapered kernel may diminish filtering artifacts and provide both accuracy and robustness benefits over baseline models.



\section{Convolution with the Hamming window}
\begin{figure}
    \begin{center}
        \includegraphics[trim= 0 0 0 0, width=0.85\linewidth]{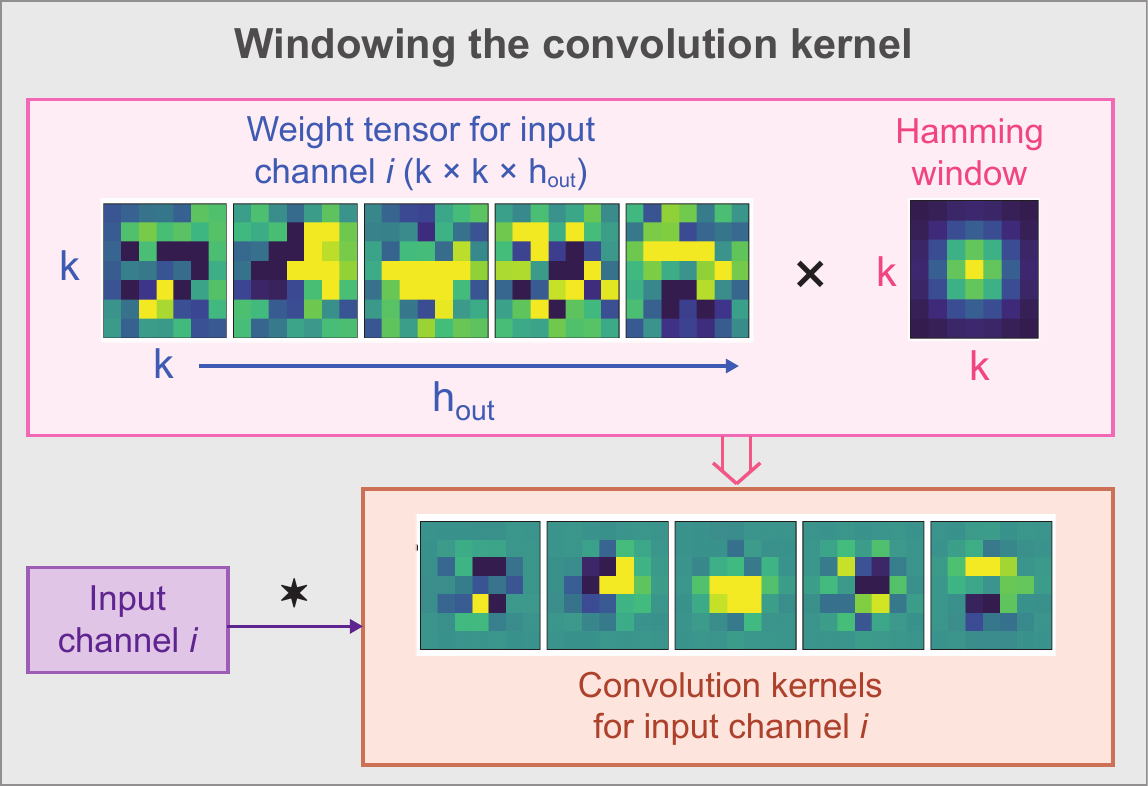}
    \end{center}
    \vspace{-0.2cm}
    \caption{
    Tapering the convolution kernels with the Hamming window. The typical weight tensor in a 2-D convolutional layer has size $(k \times k \times h_{in} \times h_{out})$. Here we show a single input channel $i$, which is convolved with $h_{out}$ distinct $k \times k$ kernels, which are generated by multiplying each $k \times k$ slice of the weight tensor with the $k \times k$ Hamming window.}
    \label{fig:windowing}
\end{figure}

\begin{figure}
    \begin{center}
        \includegraphics[trim=0 0 0 0, width=\linewidth]{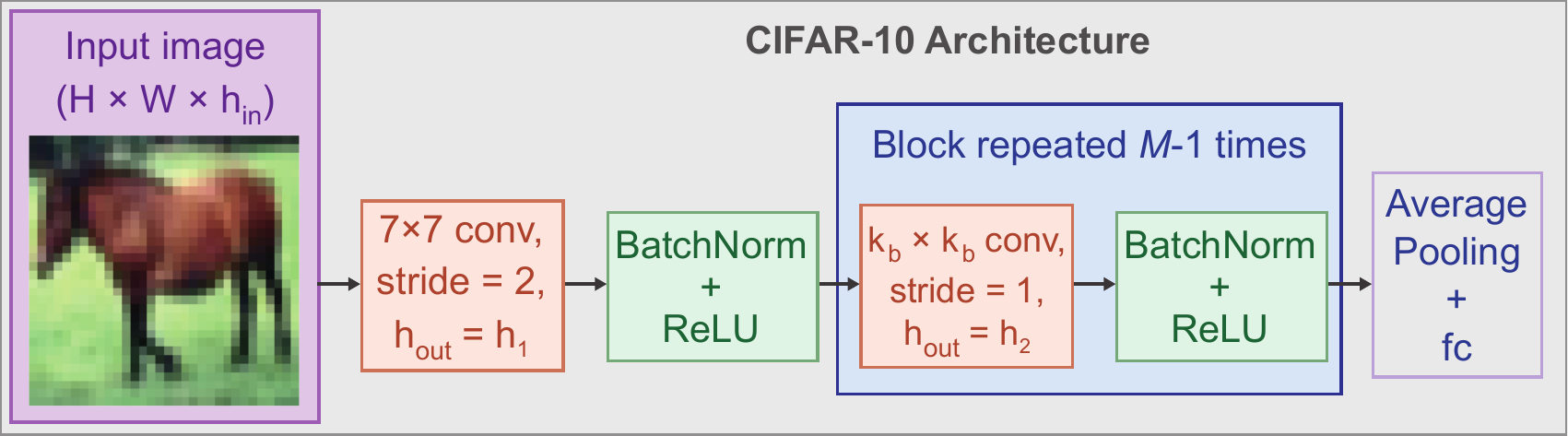}
    \end{center}
    \vspace{-0.2cm}
    \caption{Simple architecture used for CIFAR-10, CIFAR-100, Fashion-MNIST and MNIST experiments. We vary the depth (number of convolutional layers $M$) of the network by repeating a convolution block (blue). The first layer downsamples the input via a strided convolution with a $7 \times 7$ kernel, similar to ResNet architectures, while the kernel size is $k_b$ for all other convolutional layers. We also impose a channel bottleneck with the first layer having $h_1$ output channels whereas subsequent layers employ $h_2>h_1$ output channels.}
    \label{fig:architecture}
    \vspace{-0.3cm}
\end{figure}

The discrete 2-dimensional convolution operation in CNNs can be described by
\vspace{-0.15cm}
\begin{equation}\label{eq:eq_cnn_conv}
    (f * g)[x_n,y_m] = \sum_{i=1}^k \sum_{j=1}^k g[x_i,y_j] \cdot f[x_{n-i},y_{m-j}]
    \vspace{-0.15cm}
\end{equation}
where $f[x_i,y_j]$ is the (padded) input image or feature map and $g[x_i,y_j]$ is a $k \times k$ kernel. From a discrete signal processing perspective, a linear filter can be implemented via a convolution.
The aperiodic discrete convolution~\cite{Oppenheim1999} between a signal $f$ and an infinite kernel $g'$ is given by
\vspace{-0.1cm}
\begin{equation}\label{eq:eq_dc}
    (f * g')[x_n,y_m] \! \coloneqq \!\!\! \sum_{\!i=-\infty}^{\infty} \sum_{\!j=-\infty}^{\infty} \! g'[x_i,y_j] \; f[x_{n-i},y_{m-j}]
    \vspace{-0.1cm}
\end{equation}
and this formulation can be used to describe an ideal, infinite impulse response (IIR) filter. In practice, to obtain kernels of finite size, in other words for finite impulse response (FIR) filter design~\cite{Prabhu2014}, it is necessary to pick an appropriate window function $U[x_i,y_j]$ with $i,j \in \mathbb{Z}$ which limits the interval where the sum in Eq.~\ref{eq:eq_dc} is non-zero. For CNNs, this window function is a rectangle function, equivalent to simple truncation. Formally, the rectangle window function
\vspace{-0.15cm}
\begin{equation}
    U[x_i,y_j]=
    \begin{cases} 
        1, & \textrm{if } \; 1 \leq i,j \leq k \\
        0, & \textrm{else}
    \end{cases}
    \vspace{-0.05cm}
\end{equation}
multiplied with the ideal, infinite kernel $g'$ in Eq.~\ref{eq:eq_dc} as
\vspace{-0.1cm}
\begin{equation}
    \begin{split}
        (f * & g)[x_n,y_m] \! = \\
        & \sum_{i=-\infty}^{\infty} \sum_{j=-\infty}^{\infty} \!
        g'[x_i,y_j] \, U[x_i,y_j]  \; f[x_{n-i},y_{m-j}]
    \end{split}
    \vspace{-0.1cm}
\end{equation}
reduces to the CNN formulation in Eq.~\ref{eq:eq_cnn_conv}. Via the convolution theorem, multiplying $g'$ with the rectangular function $U$ in space domain corresponds to convolving the frequency response of $g'$ with a sinc function in frequency domain.
Thus, windowing via simple truncation introduces potentially unwanted frequency components into the frequency response of the finite kernel $g[x,y]$, as shown in Fig.~\ref{fig:fig1}.

As an alternative to simple truncation, we propose to reduce unwanted frequency components through the standard Hamming window~\cite{Hamming1998} in the convolution operations in a CNN. The one-dimensional Hamming window is a special case of the generalized cosine window, and is defined as
\vspace{-0.1cm}
\begin{equation}
    U[n]=\alpha-(1-\alpha) \cdot \cos\left(\frac{2\pi n}{N}\right), \;\;\; 0 \leq n \leq N
    \vspace{-0.1cm}
\end{equation}
with $\alpha=25/46$~\cite{Hamming1998,Prabhu2014} and a window size of $N$ discrete samples. We define the 2-D Hamming window as the outer product of two 1-D Hamming windows.
The Hamming window can be implemented in standard architectures simply by multiplying each two-dimensional $k \times k$ kernel in a convolutional layer with the $k\! \times \!k$ Hamming window function (Fig.~\ref{fig:windowing}).

The Hamming window can be interpreted as a form of regularization. Multiplication with the window function reduces the gradient flow, or the effective learning rate, to the kernel boundaries which keeps the boundary weights close to zero and effectively shrinks the parameter space.

\section{Experiments}

\subsection{Do CNN filters suffer from spectral leakage?}
\begin{figure}[t]
    \begin{center}
        \includegraphics[trim=0 5 0 0,clip, width=0.92\linewidth]{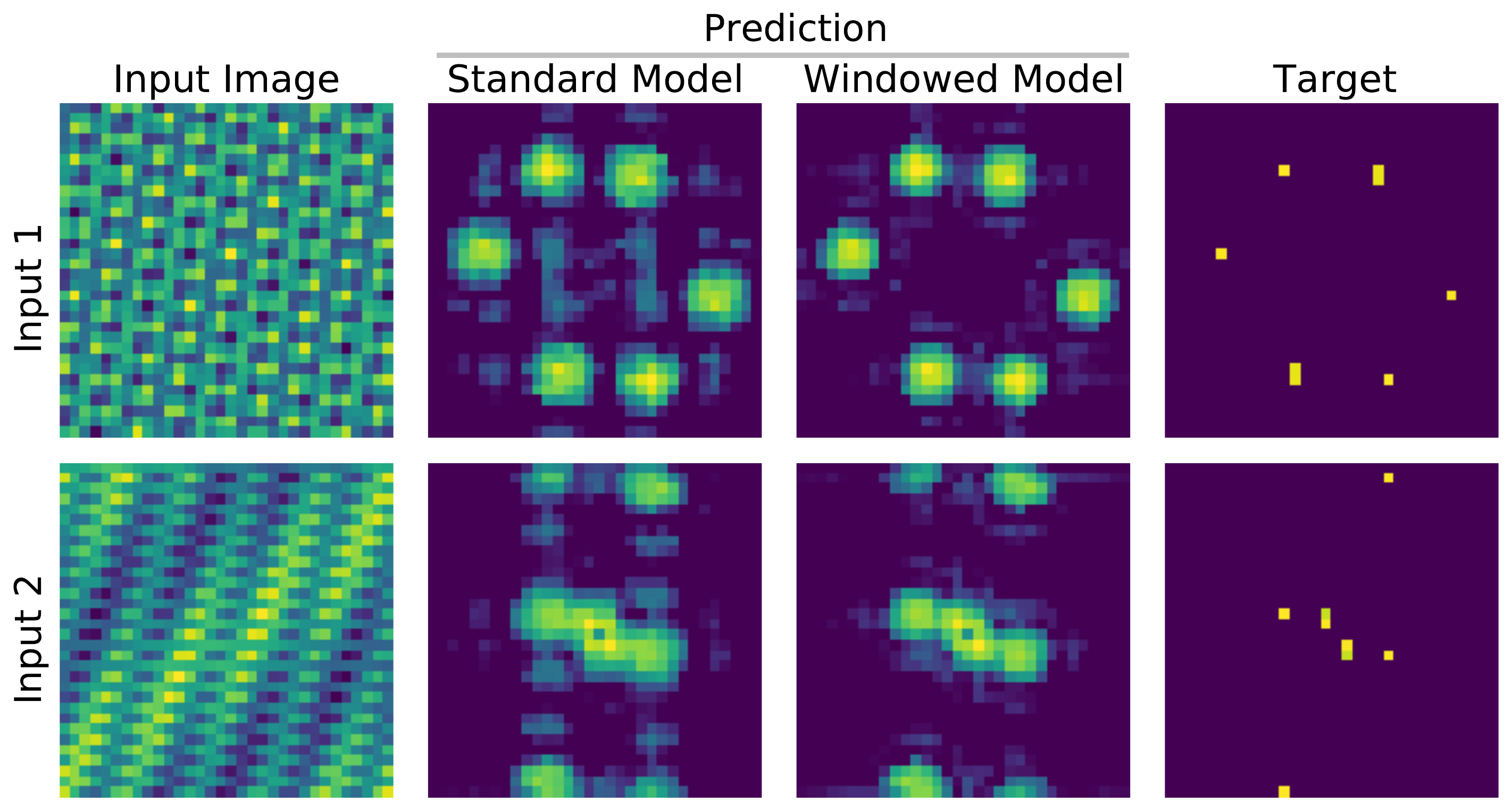}
        \caption{Learning to predict the FFT magnitude of an input image with a single convolutional layer. Network predictions for two example synthetic input images, randomly generated as the sum of three 2-D sine waves. The target vectors are the FFT magnitudes of the input images, including negative frequencies. We find that the model using Hamming windows alleviates leakage artifacts.}
        \label{fig:fftnet_pred}
    \end{center}
\end{figure}

We devise a simple, fully controlled experiment to test whether the kernels in a single convolutional layer trained in a supervised setting display spectral leakage. To address this question, we force the network to learn good-quality bandpass filters in a regression task to predict the FFT magnitude of the input image. We create a synthetic dataset where the input images $S(x,y)$ are generated randomly via the superimposition of 2-D sine waves. Each input image
\vspace{-0.25cm}
\begin{equation}
    S(x,y)=\sum_{i=1}^3 \, \sin(2 \pi \, x'_i \; \omega_i + \phi_i),
    \vspace{-0.05cm}
\end{equation}
\begin{equation}
    \textrm{with} \:\:\:\: x'_i=x \cos(\theta_i)+y \sin(\theta_i)
\end{equation}
is the sum of three 2-D sine waves with spatial frequency $\omega_i$ sampled uniformly from $[0, 0.5\omega_s)$, where $0.5\omega_s$ is the Nyquist frequency. The orientation $\theta_i$ and phase $\phi_i$ of each sine wave are also sampled uniformly in the intervals $[0, \pi)$ and $[0, 2\pi)$ respectively.
Each target (ground truth) vector is the flattened 2-D FFT magnitude of the corresponding input image. Input images are $32 \times 32$ pixels in size, hence the target values are vectors of length 1,024. Including the negative frequencies, the networks need to predict large values at 6 distinct frequency locations for each input image, as illustrated in Fig.~\ref{fig:fftnet_pred}.

\begin{figure}[t]
    \begin{center}
        \vspace{-0.4cm}
        \includegraphics[trim=0 5 5 5,clip, width=0.6\linewidth]{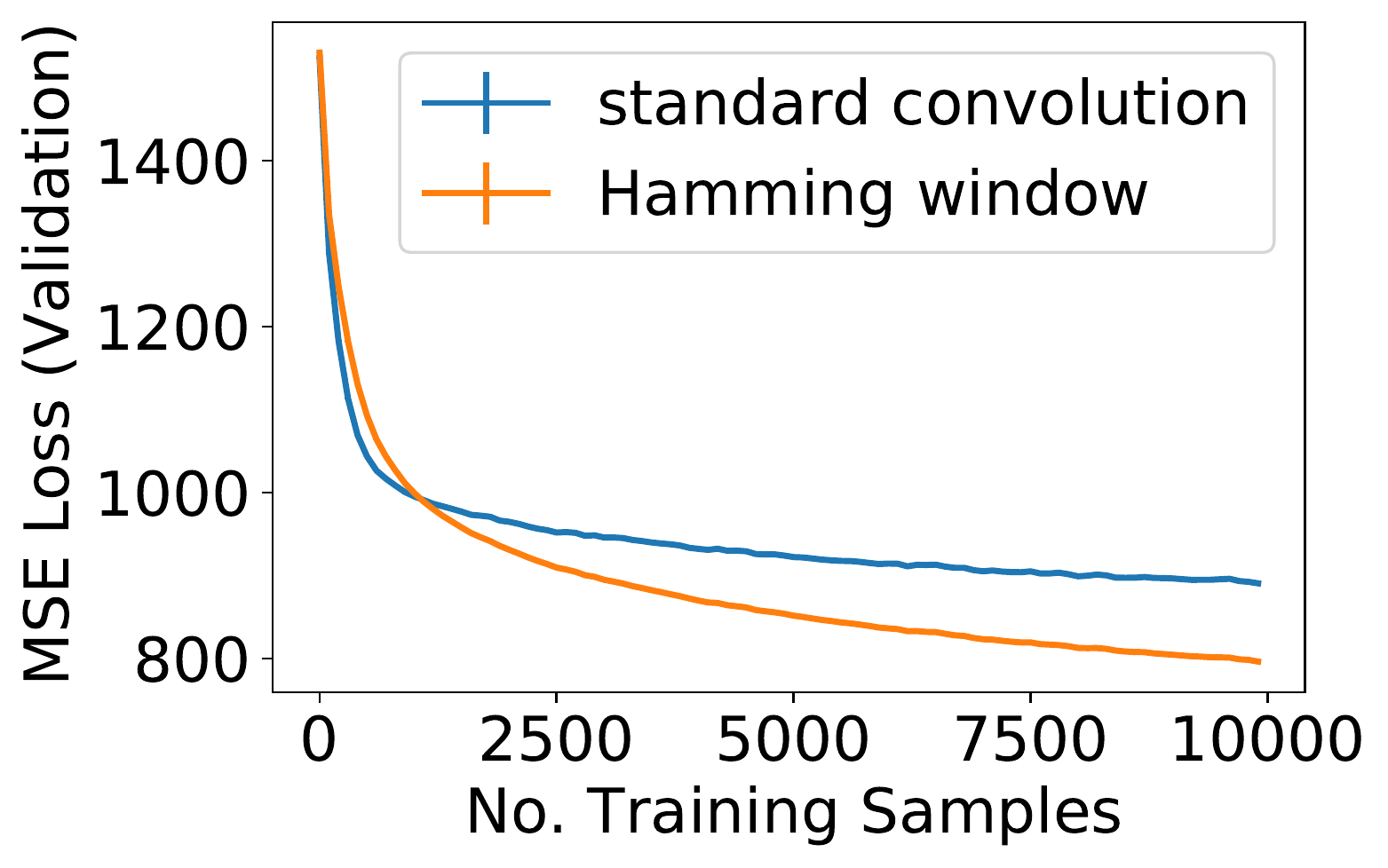}
        \caption{The windowed model achieves better regression performance on an independent validation set of 1000 images. The results are averaged over 5 runs with random model initializations (standard deviation error bars are too small to see).}
        \label{fig:fftnet_loss}
    \end{center}
    \vspace{-0.6cm}
\end{figure}
\begin{figure*}[t]
    \begin{center}
        \begin{tabular}{cc}
            \hspace{-0.4cm}
            \begin{tabular}{c}
                \includegraphics[trim=15 0 20 15,clip, width=0.66\linewidth]{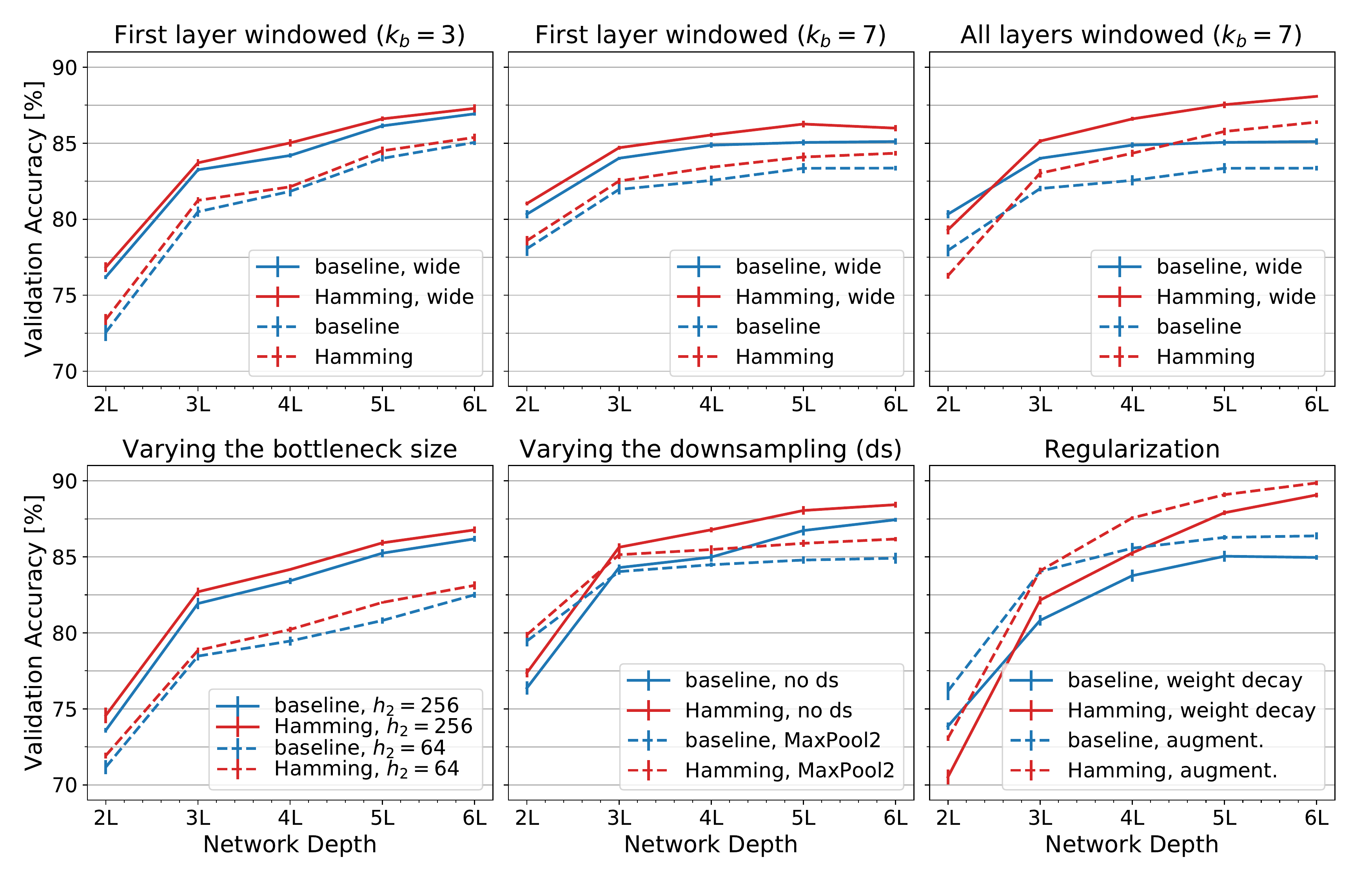}
            \end{tabular}
            &
            \hspace{-0.7cm}
            \begin{tabular}{c}
                \hspace{0.5cm}\includegraphics[trim=0 0 0 50,
                width=0.29\linewidth]{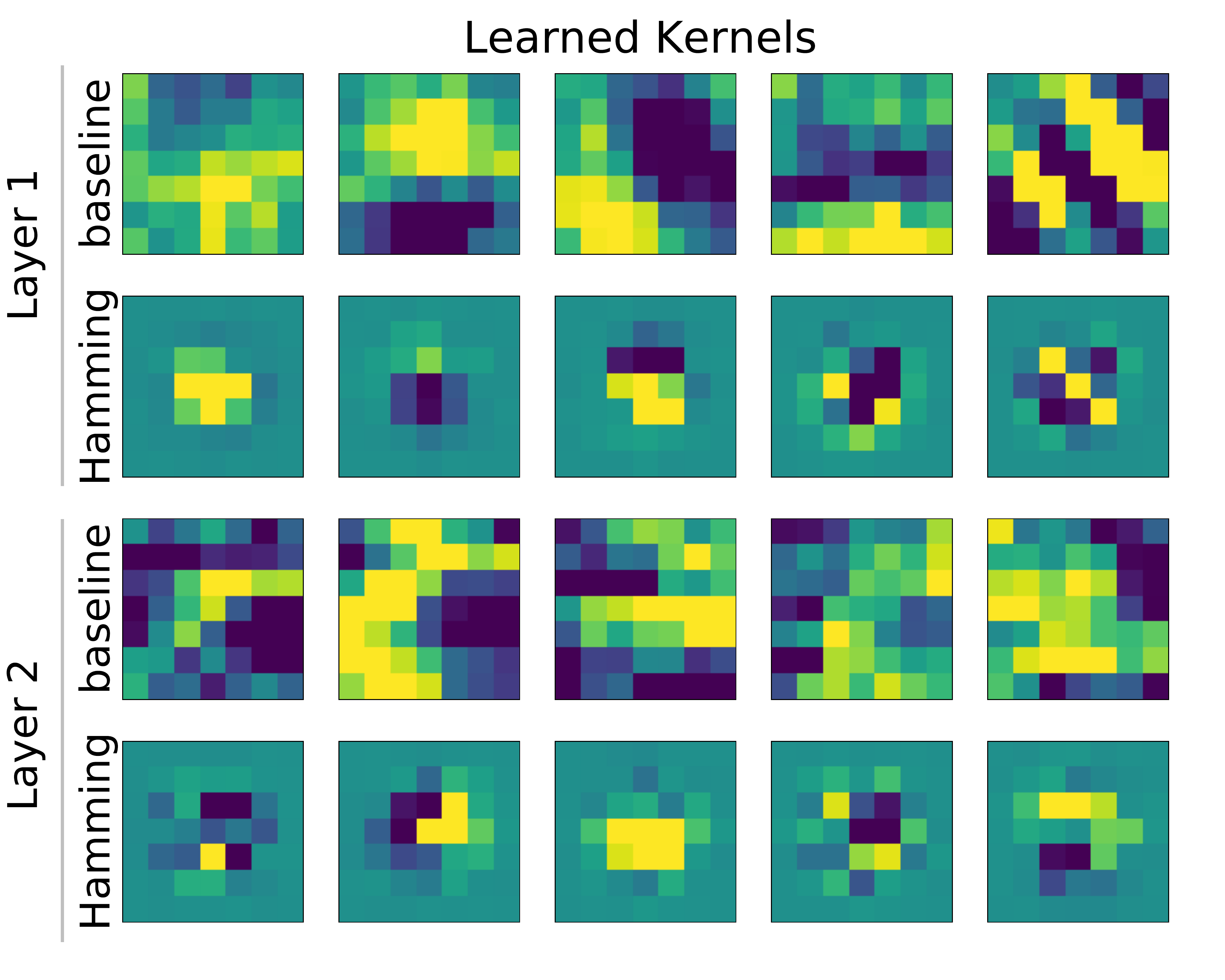} \vspace{-0.3cm}\\
                {\hspace{0.6cm} \small(b)} \\
                \includegraphics[trim=12 0 0 10,clip, width=0.31\linewidth]{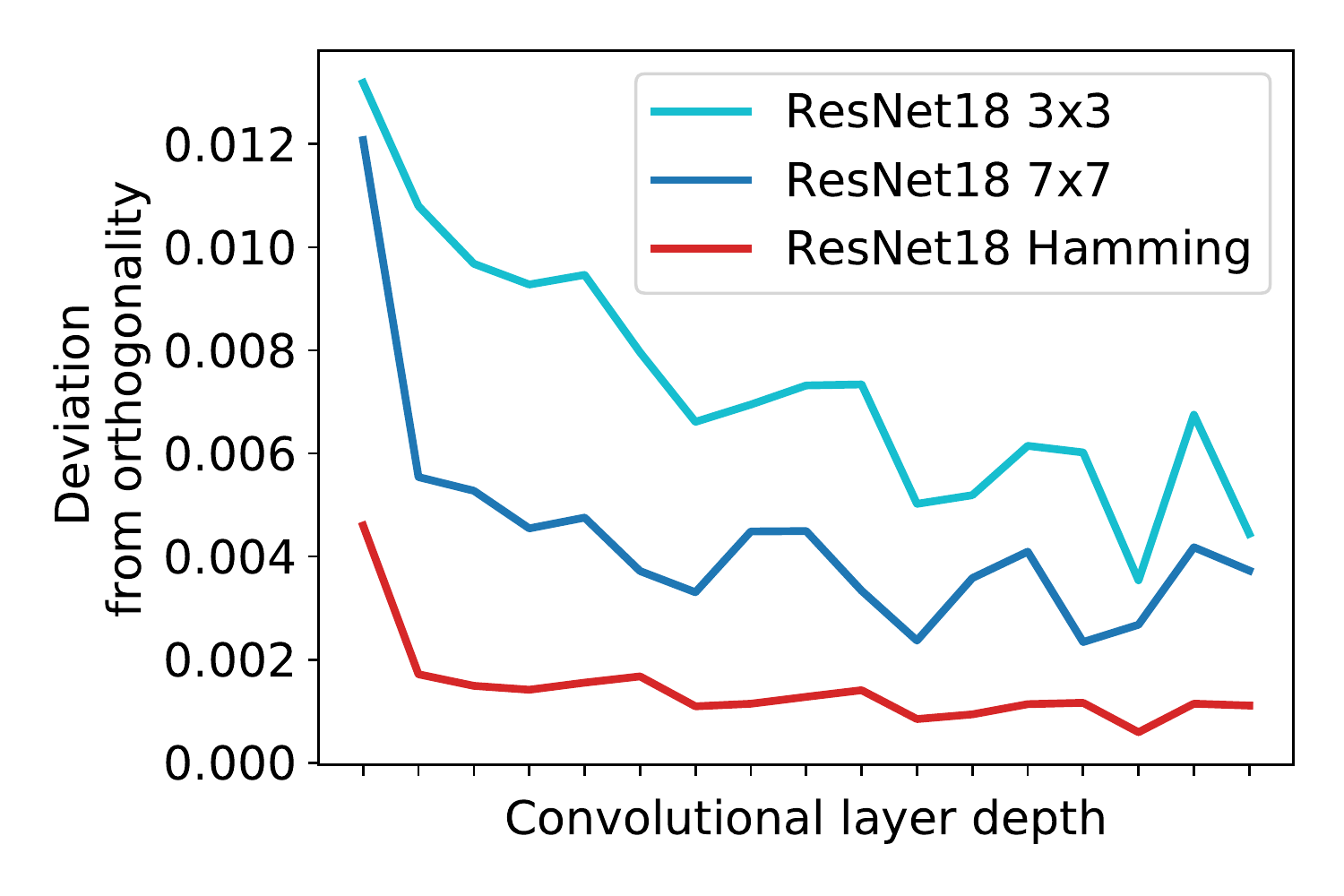}
            \end{tabular} \vspace{-0.2cm} \\
            {\small(a)} & {\small(c)}
        \end{tabular}
    \end{center}
    \vspace{-0.2cm}
    \caption{(a) CIFAR-10 validation accuracy as a function of varying network depth ($M=2\ldots6$ convolutional layers), in models using the Hamming window (red) and baselines with standard convolutional layers (blue). Line plots depict the average of 5 runs with error bars denoting standard deviation. We find that for all architecture variants we tried (Hamming window only on the first layer, Hamming window on all layers, different channel width, different methods of downsampling and regularization), models using the Hamming window consistently outperform the baseline models in networks deeper than 2 layers. (b) Example kernels after training in the network variant with $k_b=7$ where the Hamming window is applied to all layers. (c) For ResNet18 models trained on ImageNet, we find that the deviation of each convolutional layer from a row orthogonal convolution~\cite{Wang2020} is lowest for the Hamming model compared to baselines.}
    \label{fig:cifar10_perf}
    \vspace{-0.3cm}
\end{figure*}

We evaluate if a CNN can learn bandpass filters to approximate the discrete Fourier transform and use a single convolutional layer with 1,024 output channels, followed by ReLU and global average pooling. We test two CNN variants: one network with a Hamming window and one network without a Hamming window (baseline). To keep the central lobe size of the frequency responses similar between the two networks, we use a convolutional kernel size of $k{=}7$ for the baseline network and $k{=}11$ for the network with the Hamming window (see Supplement for a scan of kernel sizes).
We train both networks using the mean squared error (MSE) and ADAM~\cite{ADAM} optimizer on 10,000 training images and report the performance on an independent validation set of 1,000 images during training. 

Results in Fig.~\ref{fig:fftnet_loss} show that longer training allows the windowed network to obtain a lower regression error on the validation set. This is partly due to the increased frequency resolution of the windowed network by using a larger kernel size, and partly due to artifact reduction. By visualizing the predictions of the trained networks in Fig.~\ref{fig:fftnet_pred}, we see that the bandpass filters learned by the baseline network, with standard convolutions, indeed suffers from leakage artifacts. In comparison, the windowed model is able to suppress responses which are adequately far from the target input frequencies.

This indicates that standard CNNs are susceptible to spectral leakage and will not readily learn filters which are tapered off at the boundaries in the absence of explicit regularization, even when leakage artifacts directly contribute to the loss. We find that a standard Hamming window can be employed to regularize the kernel weights and combat leakage artifacts. However, it is not clear from this toy experiment whether deeper networks with a large number of non-linearities can learn to suppress performance-degrading artifacts. Therefore, we investigate the effects of windowing in deeper networks next.

\subsection{When does spectral leakage hurt classification?}

We extensively evaluate on CIFAR-10 and CIFAR-100 with model variations based on Fig.~\ref{fig:architecture}.

\textbf{CIFAR-10.} For all experiments, we train for 50 epochs using cross-entropy loss and SGD with a mini-batch size of 32 and momentum 0.9. The initial learning rate is 0.01 and decays by a factor of 0.1 at epochs 25 and 40.
We vary network width and depth where we evaluate from 2 layers deep up to 6 layers deep. Unless stated otherwise, the number of output channels in the convolutional layers are $h_1=32$ and $h_2=128$ for the original networks and $h_1=64$ and $h_2=256$ for `wide' networks. The windowed and baseline networks are trained identically 
and repeated 5 times with different random seeds. 

\begin{figure*}[t]
    \begin{center}
        \begin{tabularx}{\linewidth}{cc}
            \hspace{-0.3cm}
            \includegraphics[trim=20 10 40 0,clip, width=0.58\linewidth]{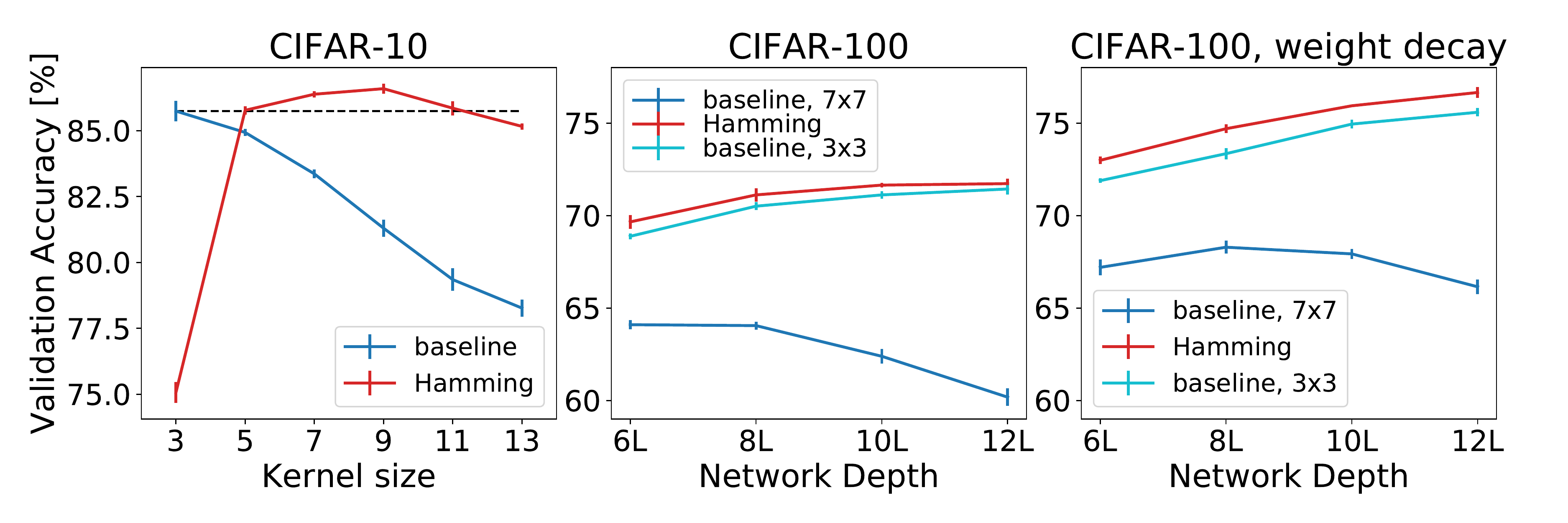} &
            \hspace{-0.4cm}
            \vspace{0.1cm}
            \multirow{3}{*}[3cm]{
            \includegraphics[trim=5 20 14 5,clip, width=0.39\linewidth]{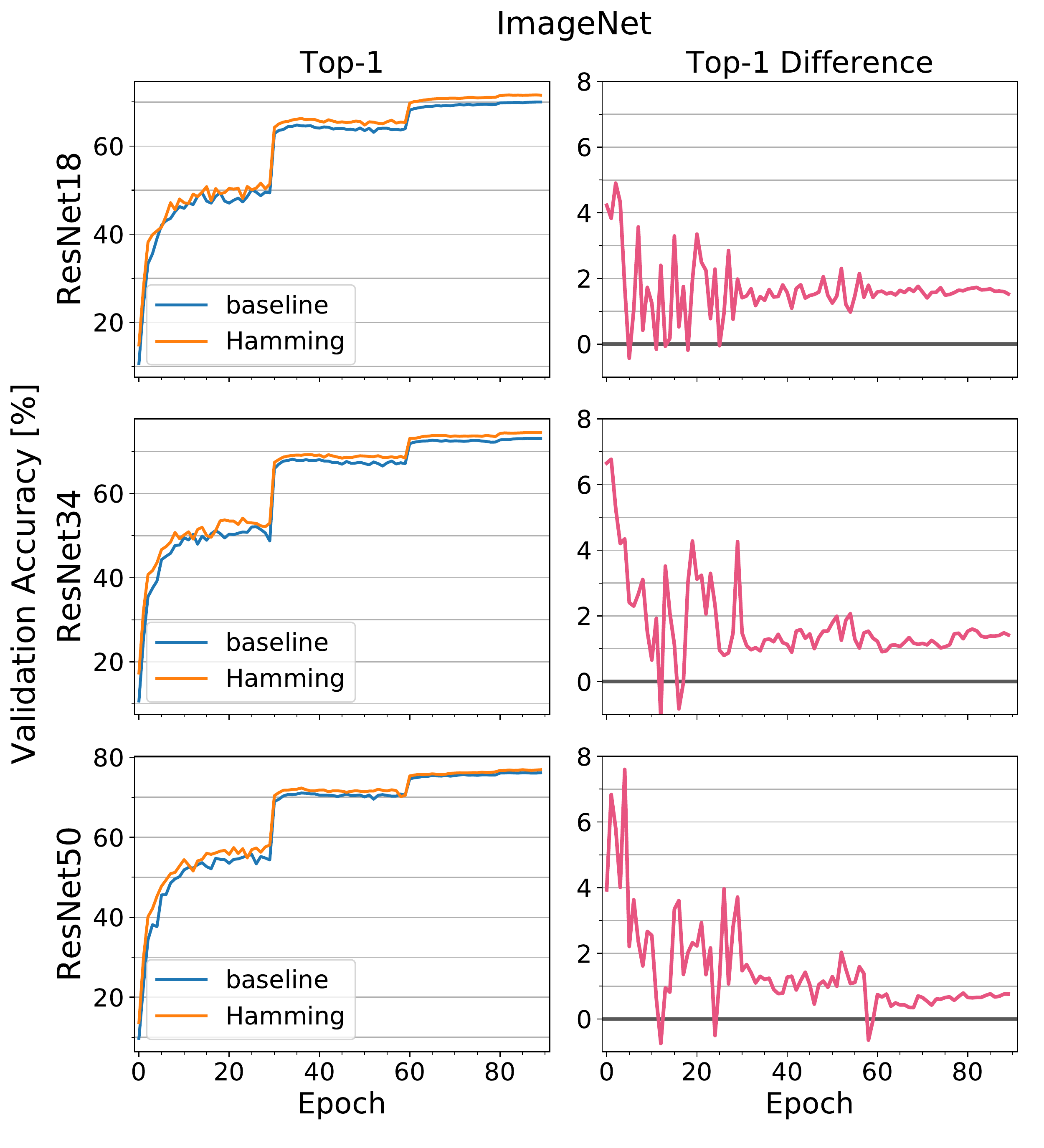}}
            \\
            {\small(a)}\\
            \hspace{-0.1cm}
            \vspace{0.1cm}
            \includegraphics[trim=15 10 14 0,clip, width=0.58\linewidth]{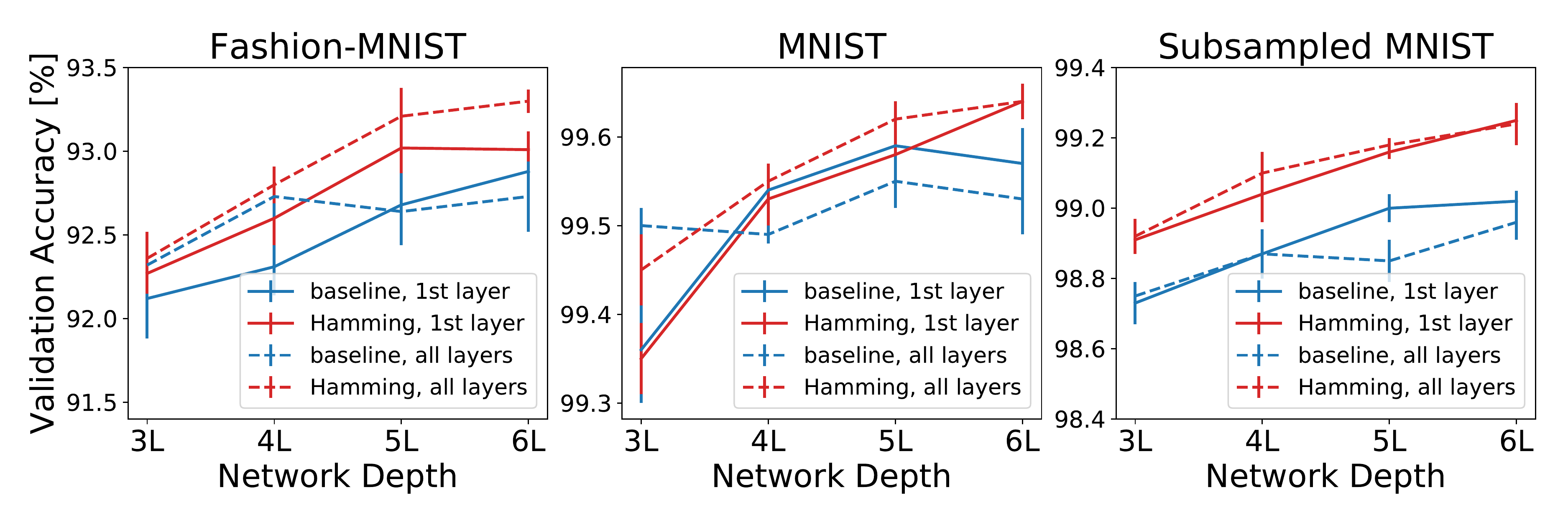}
            \vspace{-0.2cm} \\
            {\small(b)} & {\small(c)}
        \end{tabularx}
    \end{center}
    \vspace{-0.2cm}
    \caption{(a) Left: CIFAR-10 validation accuracy decreases monotonically with increasing kernel size for the baseline model, while a larger $9\times 9$ kernel size maximizes performance for the model with the Hamming window. Middle and Right: `Hamming' models with $7 \times 7$ kernels (red) outperform baseline networks with both $7 \times 7$ (blue) and $3 \times 3$ kernels (cyan) on the CIFAR-100 dataset. The performance boost is more pronounced when the networks are regularized by weight decay. (b) Left and middle: Fashion-MNIST and MNIST validation accuracy as a function of varying network depth. Right: We find that the benefits of windowing are more pronounced when the magnitude of high frequency components is increased by subsampling the input images. (c) Left: ImageNet validation accuracy is higher for ResNet architectures with the Hamming window than baseline ResNet models throughout training. Right: The difference in ImageNet validation accuracy between the Hamming and baseline ResNet models during training.
    }
    \label{fig:mnist_perf}
    \vspace{-0.3cm}
\end{figure*}

\textbf{First layer.} The earlier layers may provide sufficiently powerful and shareable features for deeper layers. Thus, we evaluate a Hamming window in only the first convolutional layer.  We test networks where the deeper layers have a kernel size of $3{\times}3$ ($k_b{=}3$) and $7{\times} 7$ ($k_b{=}7$) and find that the accuracy of windowed models is consistently higher than the baseline. This is true both for the original and wide models (Fig.~\ref{fig:cifar10_perf}a, top row). Note that the performance increase is relatively small, it is caused by a Hamming window \emph{only} in the first convolutional layer, while the rest of the architecture and hyperparameters are identical.

\textbf{All layers.} Next, we test whether alleviating spectral leakage in deeper layers. As windowing very small kernels is not meaningful, and we would like to keep the number of parameters in the baseline and windowed networks the same, we use a kernel size of $7{\times}7$ ($k_b{=}7$) in all layers. Notably, we find that using the Hamming window in all convolutional layers provides a significant boost to CIFAR-10 validation accuracy, especially in deeper networks (Fig.~\ref{fig:cifar10_perf}a, top row, right).
To illustrate the learned weights, some example kernels from trained networks are shown in Fig.~\ref{fig:cifar10_perf}b.

\textbf{Feature sharing.} We hypothesise that  artifact-free bandpass filters offers better shareable representations. Thus, the channel-wise network bottlenecks, which forces stronger feature sharing, may effect performance. To test this, we vary the bottleneck size in the original network ($h_1=32$) by changing the number of output channels in the deeper layers to $h_2=64$ and $h_2=256$, while using a Hamming window only in the first layer. Interestingly, we find that using a larger or smaller bottleneck size does not seem to affect the accuracy increase provided by the windowed convolution (Fig.~\ref{fig:cifar10_perf}a, bottom row, left).

\textbf{Aliasing.} Downsampling layers in CNNs are known to introduce performance-degrading aliasing artifacts~\cite{Zhang2019}. We investigate whether the Hamming window may also be indirectly suppressing task-irrelevant, aliased frequency components. Thus, we train networks with no downsampling layers, and replace the strided convolution with a standard convolution (stride=1) while windowing only the first layer. As another control experiment, we also train networks which perform downsampling via a max-pooling layer with a $2{\times}2$ window instead of a strided convolution. We find that in both cases using a Hamming window still improves CIFAR-10 validation accuracy (Fig.~\ref{fig:cifar10_perf}a, bottom row, middle), which indicates that the performance increase provided by windowed convolutions is independent of aliasing and the choice of downsampling method.

\textbf{Regularization.} Our windowing regularizes the kernel weights close to the boundaries (Fig.~\ref{fig:cifar10_perf}b).  We compare this regularization with other common methods, namely weight decay and data augmentation. We train networks with a weight decay value of 0.001, and independently, we train networks with random translation and horizontal flip augmentations. We find that, in explicitly regularized networks, not only do the benefits of our windowing not disappear, but the accuracy boost is in fact larger, especially for deeper networks (Fig.~\ref{fig:cifar10_perf}a, bottom row, right).

\indent \textbf{Optimal kernel size.} Standard convolutional layers typically employ very small ($3{\times}3$) kernel sizes. Although we only used a kernel size of $7{\times}7$ so far for our proposed method, it is not clear \emph{a priori} what kernel size would be optimal for the Hamming window. To test this empirically, we vary the kernel size in all layers of a $M{=}6$ layer network, with or without the Hamming window. We find that while classification performance decreases monotonically with increasing kernel size beyond $3{\times }3$ for the baseline network, there is a larger, optimal kernel size which maximizes performance for the network using the Hamming window (Fig.~\ref{fig:mnist_perf}a, left). In fact, we find that windowed networks with both kernel sizes $k{=}7$ and $k{=}9$ provide a significant accuracy improvement (outside of the standard deviation) over the best baseline model with $k{=}3$.

\textbf{CIFAR-100.} For CIFAR-100~\cite{cifar10} experiments we employ wider models with $h_1{=}128$ and $h_2{=}256$, deeper models with up to 12 layers, and for the `Hamming' models we use windowed convolutions in all layers. We train all models for 150 epochs, with initial learning rate 0.01 decaying by a factor of 0.1 at epochs 75 and 120. We also employ standard data augmentation (horizontal flip and random translations). Otherwise, the hyperparameters are the same as in the CIFAR-10 experiments. As an additional control, we run baselines with $7{\times} 7$ kernel size (same number of parameters as the `Hamming' model) and $3 {\times} 3$ kernel size (best baseline performance) in all layers. We find that windowed networks perform consistently better than both baselines (Fig.~\ref{fig:mnist_perf}a, middle). The accuracy enhancement provided by the Hamming window is more pronounced with a weight decay of 0.001 (Fig.~\ref{fig:mnist_perf}a, right).

\vspace{-0.03cm}
\subsection{Datasets with limited frequencies}
\vspace{-0.07cm}
Natural images may contain class-specific information in all frequency bands, which means spectral leakage between different frequency components may hinder discrimination of class-specific responses.
We hypothesize that for less natural images, where not all frequency components are well-represented in the training set, the effects of windowing would be less prominent. To test this, we evaluate classification performance on Fashion-MNIST~\cite{FMNIST} and MNIST~\cite{MNIST} datasets.
Training parameters in this section are identical to the CIFAR-10 experiments. We train two types of models: one with convolutions with the Hamming window only in the first layer and $k_b=3$ and one with Hamming window in all layers with $k_b=7$. 

For the Fashion-MNIST dataset, we find that the use of a Hamming window persistently improves classification performance (Fig.~\ref{fig:mnist_perf}b, left), however the increase in validation accuracy is smaller than it is with more natural images found in CIFAR-10 and CIFAR-100 datasets. For the MNIST dataset, we don't observe a performance increase in `Hamming' models for most networks, and only a modest one for deeper networks (Fig.~\ref{fig:mnist_perf}b, middle).

We attribute the lack of benefits from windowed convolutions in the MNIST dataset, to some degree, to the lack of high frequency components, whereby leakage in lowpass and bandpass filters cannot contaminate high frequency information. To test this, we subsample the $28 \times 28$ input images in the MNIST dataset, via bilinear interpolation, down to $14 \times 14$ images. Subsampling has the effect of increasing the relative magnitude of high frequency components, and we find that the Hamming window provides significant accuracy improvements in the subsampled MNIST (Fig.~\ref{fig:mnist_perf}b, right). In particular, we find that both `Hamming' models (windowing only the first layer or all layers) perform better than both baselines (including when $k_b=3$).

\vspace{-0.05cm}
\subsection{ImageNet}
\vspace{-0.05cm}
We train ResNet~\cite{He2015} and VGG~\cite{Simonyan2015} models on the ImageNet~\cite{Imagenet} dataset for 90 epochs, with initial learning rate of 0.1 decaying by 0.1 at epochs 30 and 60. Optimization is performed using SGD with momentum 0.9 and weight decay $10^{-4}$. Input images are randomly resized and cropped to $224 \times 224$ pixels and horizontally flipped.
Baseline networks are VGG architectures with batch normalization~\cite{Ioffe2015} and kernel size $k=3$ or $k=7$, and the standard ResNet architectures~\cite{He2015} with $k=7$ in the first layer, and $k=3$ or $k=7$ in all deeper layers. For the windowed networks, we replace all convolutions with Hamming-windowed convolutions with $k=7$.

\begin{figure}[t]
    \vspace{-0.15cm}
    \centering
    \includegraphics[width=0.82\linewidth]{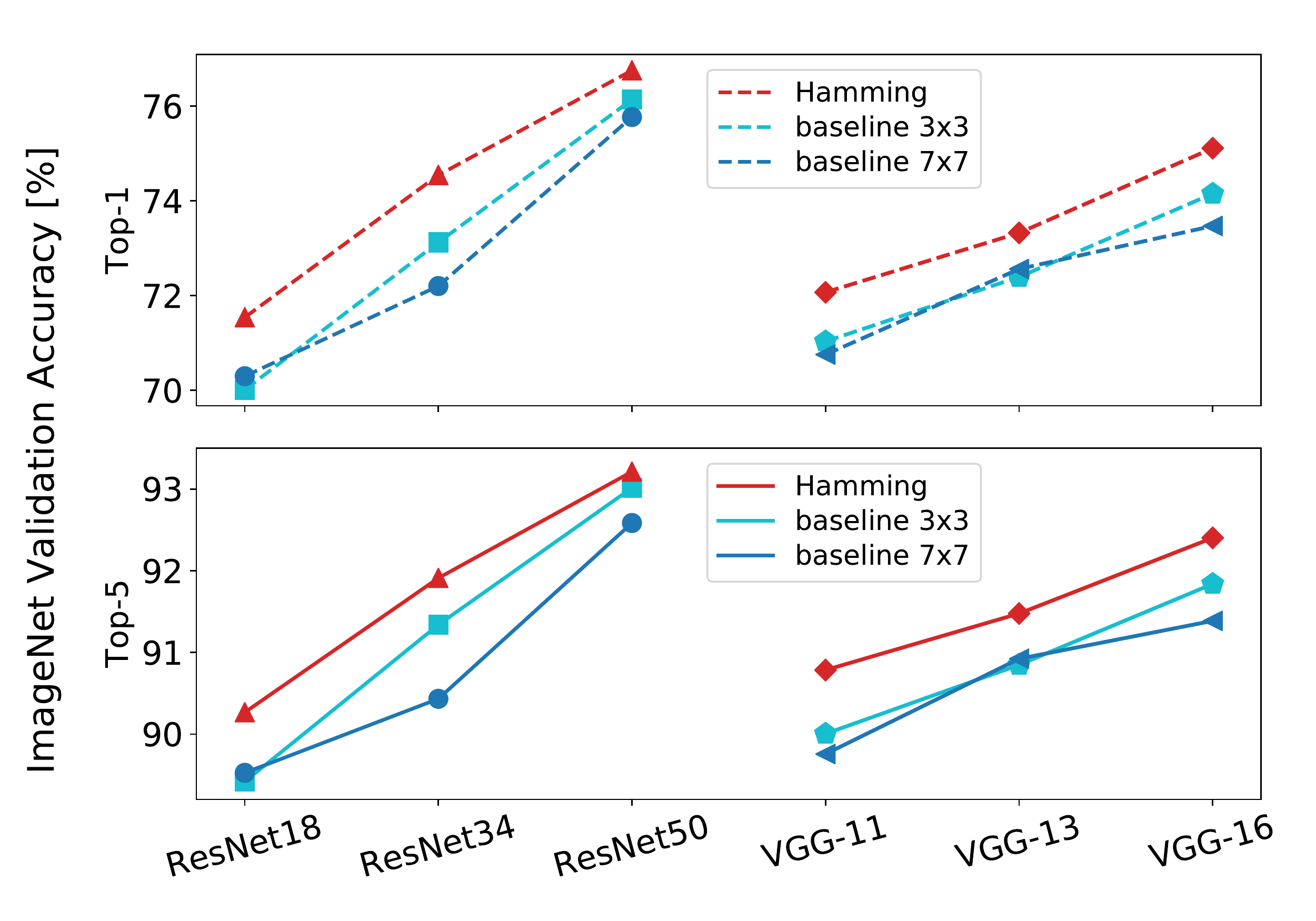}
    \caption{ImageNet validation accuracy for the baseline ResNet and VGG models and their windowed counterparts, where all kernels are replaced with $7\times 7$ Hamming windowed kernels.}
    \label{fig:imagenet_scatter}
    \vspace{-0.2cm}
\end{figure}

We find that when enforcing better frequency-selectivity, ImageNet validation accuracy is higher than baselines throughout training (Fig.~\ref{fig:mnist_perf}c).
This indicates that using standard window functions as an inductive prior helps the network settle early on to solutions which generalize better.
Overall, we find that replacing the convolutional layers with Hamming-windowed layers provides accuracy improvements on the ImageNet benchmark (Fig.~\ref{fig:imagenet_scatter}, Table~\ref{tb:imagenet_accuracy}).

\textbf{Orthogonality.} Spectral leakage may render CNNs unable to learn filters with non-overlapping frequency responses, thus leading to redundant representations. Similar to our windowed layers, redundancy reduction and performance increase can also be achieved by orthogonal convolutions~\cite{Wang2020}. Therefore, we analyse the effects of the Hamming window on the orthogonality of the weights learned by the ResNet18 model. A row orthogonal convolution can be written as a matrix multiplication $\mathbf{y}=\mathcal{K}\mathbf{x}$ of the input $\mathbf{x} \in \mathbb{R}^{CHW}$ with the doubly block-Toeplitz matrix $\mathcal{K} \in \mathbb{R}^{(MH'W') \times (CHW)}$ with the orthogonality condition
\vspace{-0.3cm}
\begin{equation}\label{eq:orth}
    \langle \mathcal{K}_{i,\boldsymbol{\cdot}},\mathcal{K}_{j,\boldsymbol{\cdot}} \rangle=
    \begin{cases} 
        1, & \textrm{if } \; i=j \\
        0, & \textrm{else}
    \end{cases}
\vspace{-0.3cm}
\end{equation}
where $C$ and $M$ denote the input and output channels, $H$ and $W$ ($H'$ and $W'$) the spatial dimensions of input $\mathbf{x}$ (output $\mathbf{y}$), and $i$ and $j$ are row indices of $\mathcal{K}$. For our ResNet18 models trained on ImageNet, we compute the pairwise dot product in Eq.~\ref{eq:orth} between every row of $\mathcal{K}$ in each layer, and present its mean deviation from the orthogonality condition in Fig.~\ref{fig:cifar10_perf}c. We find that the convolution operators deviate from orthogonality the least when using the Hamming window. We suggest that enforcing orthogonality may be one explanation for the performance increase displayed by windowed convolutions.
(See Supplement for further analysis on different models.)

\subsection{Adversarial attacks}
We test the robustness of baseline and Hamming models (with $M=6$ layers) trained on the CIFAR-10 dataset against DeepFool~\cite{Deepfool} (white-box) and spatial transformation~\cite{Engstrom2019} (black-box) attacks.
DeepFool attacks are iterative attacks designed to minimize the norm of the perturbation while generating examples fast, which we believe is an effective method. Similarly, spatial transformations provide a realistic black-box setting.
We generate attacks using the Adversarial Robustness Toolbox~\cite{ART2018}. For the DeepFool attacks we use 100 maximum iterations, and for the spatial transformations we use different maximum translation and rotation values as given in Table~\ref{tb:adversarial}b.

\begin{table}
    \begin{center}
        \small
        \begin{tabular}{lcc}
            Model & Top-1 (\%) & Top-5 (\%) \\
            \toprule
            ResNet18 & 70.01 & 89.42 \\
            ResNet18 $7\times 7$ & {70.30} & {89.53} \\
            ResNet18 + Hamming $7\times 7$ & \textbf{71.54} & \textbf{90.27} \\
            \midrule
            ResNet34 & 73.12 & 91.34 \\
            ResNet34 $7\times 7$ & {72.20} & {90.43} \\
            ResNet34 + Hamming $7\times 7$ & \textbf{74.54} & \textbf{91.91} \\
            \midrule
            ResNet50 & 76.14 & 93.01 \\
            ResNet50 $7\times 7$ & {75.77} & {92.58} \\
            ResNet50 + Hamming $7\times 7$ & \textbf{76.80} & \textbf{93.21} \\
            \toprule
            VGG-11 & 71.03 & 90.00 \\
            VGG-11 $7\times 7$ & {70.75} & {89.76} \\
            VGG-11 + Hamming $7\times 7$ & \textbf{72.07} & \textbf{90.78} \\
            \midrule
            VGG-13 & 72.39 & 90.85 \\
            VGG-13 $7\times 7$ & {72.56} & {90.92} \\
            VGG-13 + Hamming $7\times 7$ & \textbf{73.32} & \textbf{91.48} \\
            \midrule
            VGG-16 & 74.15 & 91.84 \\
            VGG-16 $7\times 7$ & {73.47} & {91.39} \\
            VGG-16 + Hamming $7\times 7$ & \textbf{75.11} & \textbf{92.40} \\
            \bottomrule
        \end{tabular}
    \end{center}
    \caption{ImageNet validation accuracies in Fig.~\ref{fig:imagenet_scatter}.}
    \label{tb:imagenet_accuracy}
    \vspace{-0.5cm}
\end{table}

We compare the validation accuracy of baseline and Hamming models with equal number of parameters on the adversarially perturbed CIFAR-10 validation set. For DeepFool attacks, we find that Hamming models with $7\times 7$ kernel size provides the best robustness in terms of the decrease in validation accuracy under perturbation (Table~\ref{tb:adversarial}a). With $5\times 5$ kernels, Hamming models perform worse under DeepFool attacks than baselines, even though the base accuracy on clean samples is higher for Hamming models. For larger kernel sizes, however, the robustness of Hamming models is significantly better.
For spatial transform attacks, we find a similar pattern. While validation accuracy decreases across the board for increasing perturbation magnitude, Hamming models with $7\times 7$ and $9\times 9$ kernels are always significantly more robust than the baseline models (Table~\ref{tb:adversarial}b).
\begin{table}
    \begin{center}
        \small
        \begin{tabularx}{\linewidth}{cccc}
            \multicolumn{4}{c}{DeepFool - Validation Accuracy (\%)}\\
            \toprule
            \multirow{2}{*}{Model} &
            \multicolumn{3}{c}{Kernel Size} \\
            \cline{2-4}
            {} & \mbox{$5\times 5$} & $7\times 7$ & $9\times 9$ \\
            \toprule
            baseline & \textbf{24.85}$\pm$0.34 & 20.06$\pm$0.13 & 18.24$\pm$0.44 \\
            Hamming & 23.20$\pm$0.29 & \textbf{32.64}$\pm$0.39 & \textbf{27.88}$\pm$0.94 \\
            \midrule
            baseline-clean & 84.93$\pm$0.13 & 83.36$\pm$0.16 & 81.30$\pm$0.32 \\
            Hamming-clean & \textbf{85.77}$\pm$0.16 & \textbf{86.38}$\pm$0.12 & \textbf{86.59}$\pm$0.19 \\
            \bottomrule
            \multicolumn{4}{c}{\small (a)}
        \end{tabularx}
    \end{center}
    \vspace{-0.5cm}
    \begin{center}
        \small
        \begin{tabular}{c|cc|cc}
            \multicolumn{5}{c}{Spatial Transformation - Validation Accuracy (\%)}\\
            \toprule
            \multirow{2}{*}{Model} & \multicolumn{2}{c}{Params} & 
            \multicolumn{2}{c}{Kernel Size} \\
            \cline{2-3}
            \cline{4-5}
            {} & {tr} & {rot} & $7\times 7$ & $9\times 9$ \\
            \toprule
            baseline & \multirow{2}{*}{12.5} &\multirow{2}{*}{22.5} &  44.59$\pm$3.12 & 38.74$\pm$1.37 \\
            Hamming & {} & {} & \textbf{53.03}$\pm$1.91 & \textbf{52.22}$\pm$1.64 \\
            \midrule
            baseline & \multirow{2}{*}{25.0} & \multirow{2}{*}{22.5} & 31.61$\pm$2.87 & 27.26$\pm$1.05 \\
            Hamming & {} & {} & \textbf{41.44}$\pm$2.87 & \textbf{38.67}$\pm$1.46 \\
            \midrule
            baseline & \multirow{2}{*}{25.0} & \multirow{2}{*}{45.0} & 19.47$\pm$0.67 & 18.13$\pm$1.21 \\
            Hamming & {} & {} & \textbf{26.42}$\pm$1.13 & \textbf{24.55}$\pm$1.87 \\
            \bottomrule
            \multicolumn{5}{c}{\small (b)}
        \end{tabular}
    \end{center}
    \vspace{-0.2cm}
    \caption{Adversarial robustness in baseline and Hamming models. All results are averaged over 5 runs. (a) Classification accuracy on the CIFAR-10 validation set with and without (clean) perturbations created by the DeepFool attack. (b) Classification accuracy on the CIFAR-10 validation set with spatial transformation attacks for different maximum translation (tr) and rotation (rot) values. Accuracy for unperturbed images is the same as in (a).}
    \label{tb:adversarial}
    \vspace{-0.3cm}
\end{table}
\vspace{-0.15cm}
\section{Conclusion}
\vspace{-0.05cm}
We investigate the impact of spectral leakage in the context of CNNs and show that convolutional layers employing small kernel sizes may be susceptible to performance-degrading leakage artifact. As a solution, we propose the use of a standard Hamming window on larger kernels, in line with well-known principles of filter design. We demonstrate enhanced classification accuracy on benchmark datasets, in models with the Hamming window. Finally, we show improved robustness against DeepFool and spatial transformation attacks in windowed CNNs.

This work is based on a simple and well-studied idea, which provides practical benefits in deep networks, highlighting the importance of signal processing fundamentals. We believe our work opens up new research questions regarding other principles of filter design. We show that the use of larger kernels, which are computationally more expensive, but parallelizable compared to deeper networks, is a viable option when the kernels are windowed properly. Computational complexity of large kernels may also be reduced via spatial factorization~\cite{Szegedy2016}.
We note that window functions may provide benefits in domains outside of computer vision, such as audio processing, where larger kernel sizes are common.

{\small
\bibliographystyle{ieee_fullname}
\bibliography{egbib}
}
\newpage
\clearpage
\newpage


\renewcommand\thefigure{S.\arabic{figure}}
\renewcommand\theequation{S.\arabic{equation}}
\renewcommand\thesection{\Alph{section}}

\setcounter{figure}{0}
\setcounter{equation}{0}
\setcounter{section}{0}

\graphicspath{{supp_figures/}}
\onecolumn
\centering

{\LARGE \textbf{Spectral Leakage and Rethinking the Kernel Size in} CNNs} \\
\vspace{0.5cm}
{\LARGE \textbf{-- Supplementary material --}}
\vspace{0.3cm}
\justify
\section{Frequency resolution and scan of kernel sizes for the FFT regression task}
As mentioned in Section 2.1, there are trade-offs between different window functions which should be taken into account in filter design. One of the main differences between using a conventional CNN kernel with the rectangular window, and using a standard tapering function is that a tapering function will effectively limit the size of the kernel in space domain. For example, once we taper a $7 \times 7$ kernel with the application of a standard window function, such as a Hamming window, the values of the weights close to the boundaries will be strongly attenuated, reducing the effective kernel size to a value more similar to $5 \times 5$ or $3 \times 3$ as shown in Fig. 6b. For a bandpass filter, the reduction of the window size in space domain leads to an increase in the width of the passband (or central lobe) in frequency domain via the uncertainty principle. This fundamentally decreases the frequency resolution attainable by the network for decreasing kernel size.

\begin{figure}[h]
    \begin{center}
        \includegraphics[trim=0 40 0 0, width=0.67\linewidth]{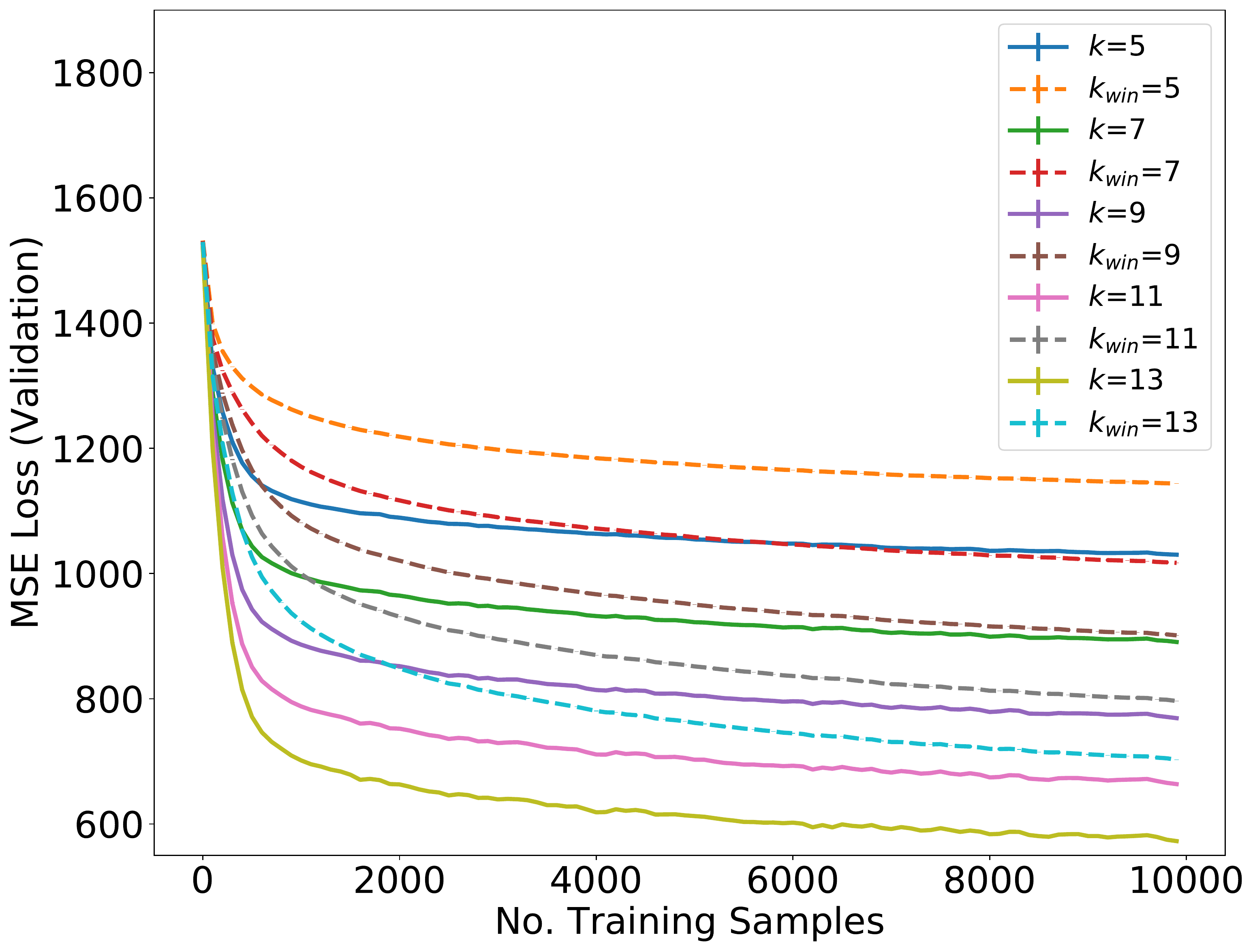}
    \end{center}
    \caption{Mean squared error loss in the baseline (solid lines) and Hamming (dashed lines) models with different kernel sizes on the FFT regression task. The loss is computed for an independent validation set of 1000 images. $k$ and $k_{win}$ refer to the kernel size in the baseline and Hamming models respectively. The setup of the experiments are identical to those described in Section 4.1 and to the learning curves shown in Fig. 5.
    The loss is averaged over 3 runs with random model initializations (standard deviation error bars are too small to see).
    }
    \label{fig:supp_loss}
\end{figure}

This effect is best demonstrated in the toy setup (Section 4.1) of the FFT regression experiment. In this setting, the minimum attainable regression loss (MSE loss) is limited by the maximum frequency resolution achievable by the network, which is determined by the kernel size. For the results presented in Section 4.1 (and shown in Fig. 4 and 5) we used a $7 \times 7$ kernel in the conventional convolutional layers and an $11 \times 11$ kernel in the layers using the Hamming window, in order to have a comparable frequency resolution between the two networks. However, it is not trivial to exactly match the frequency resolutions of such small, discrete kernels, and therefore, not easy to disentangle how much of the better performance in the Hamming model can be attributed to the frequency resolution and how much to artifact suppression. To survey the effect of the different kernel sizes, we perform a parameter scan of the kernel size in both networks. The resulting validation learning curves are shown in Fig.~\ref{fig:supp_loss}.

\begin{figure}[t]
    \begin{center}
        \includegraphics[trim=0 0 0 0,clip, width=0.99\linewidth]{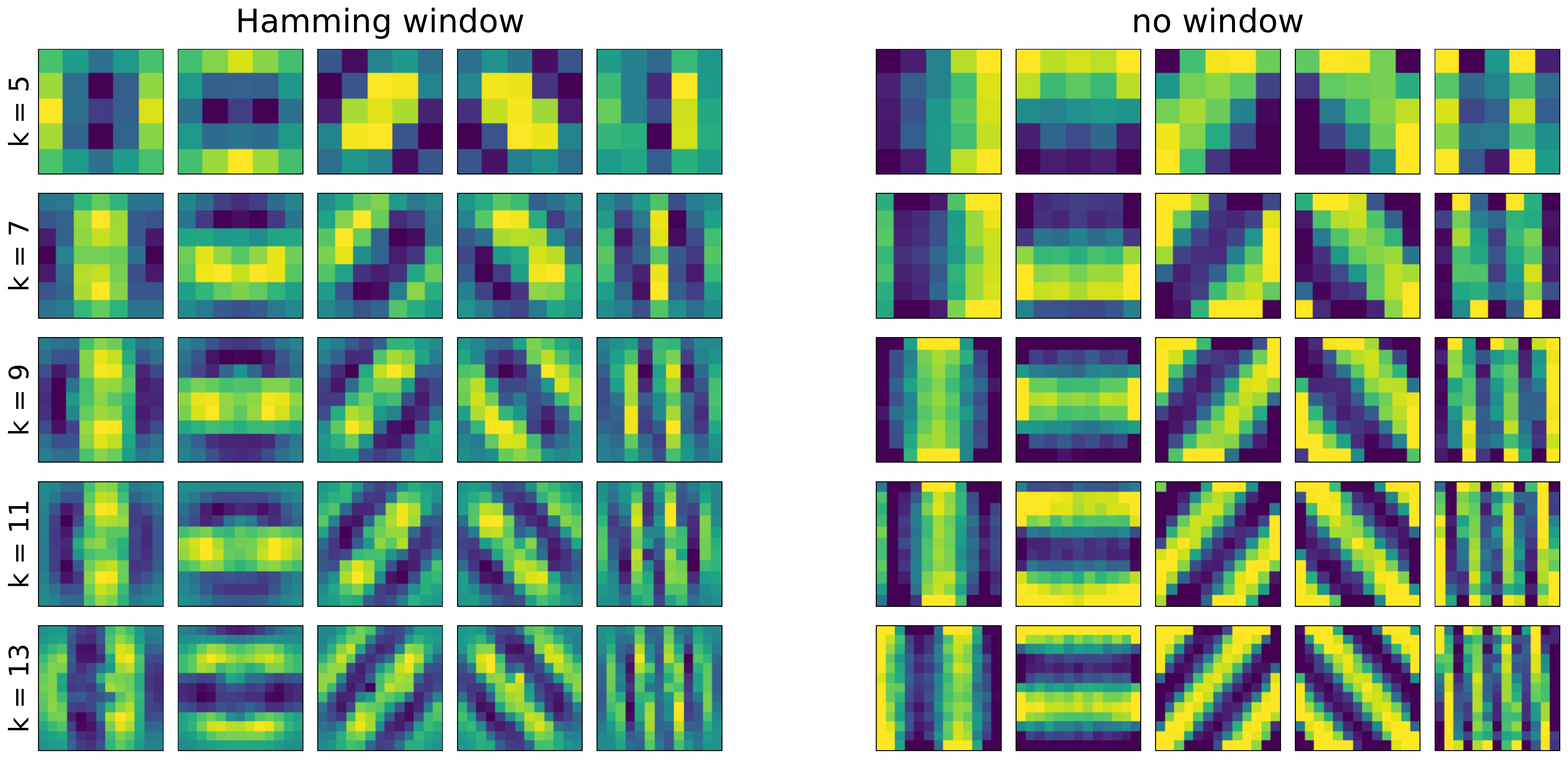}
    \end{center}
    \caption{Example filters learned in the FFT regression task by the Hamming models (left) and baseline models (right) with different kernel sizes $k$ (rows). Each column corresponds to the kernel in one output channel of the convolutional layer, or equivalently, one frequency bin of the target FFT. We find that similar bandpass filters are learned in both models, however, the kernels are tapered of in space domain for the Hamming models.}
    \label{fig:supp_filters}
\end{figure}

To give some insight into the effects the Hamming window has on the learned filters, we plot the kernels corresponding to different FFT frequency bins in both baseline and Hamming models after training (Fig.~\ref{fig:supp_filters}). It is clear that very similar bandpass filters can be learned by both models, given large enough kernel sizes, however, the multiplication with a Hamming window in space domain has a tapering effect on the kernels in the Hamming models.

\begin{figure}[t]
    \begin{center}
        \includegraphics[trim=0 0 0 0,clip, width=0.99\linewidth]{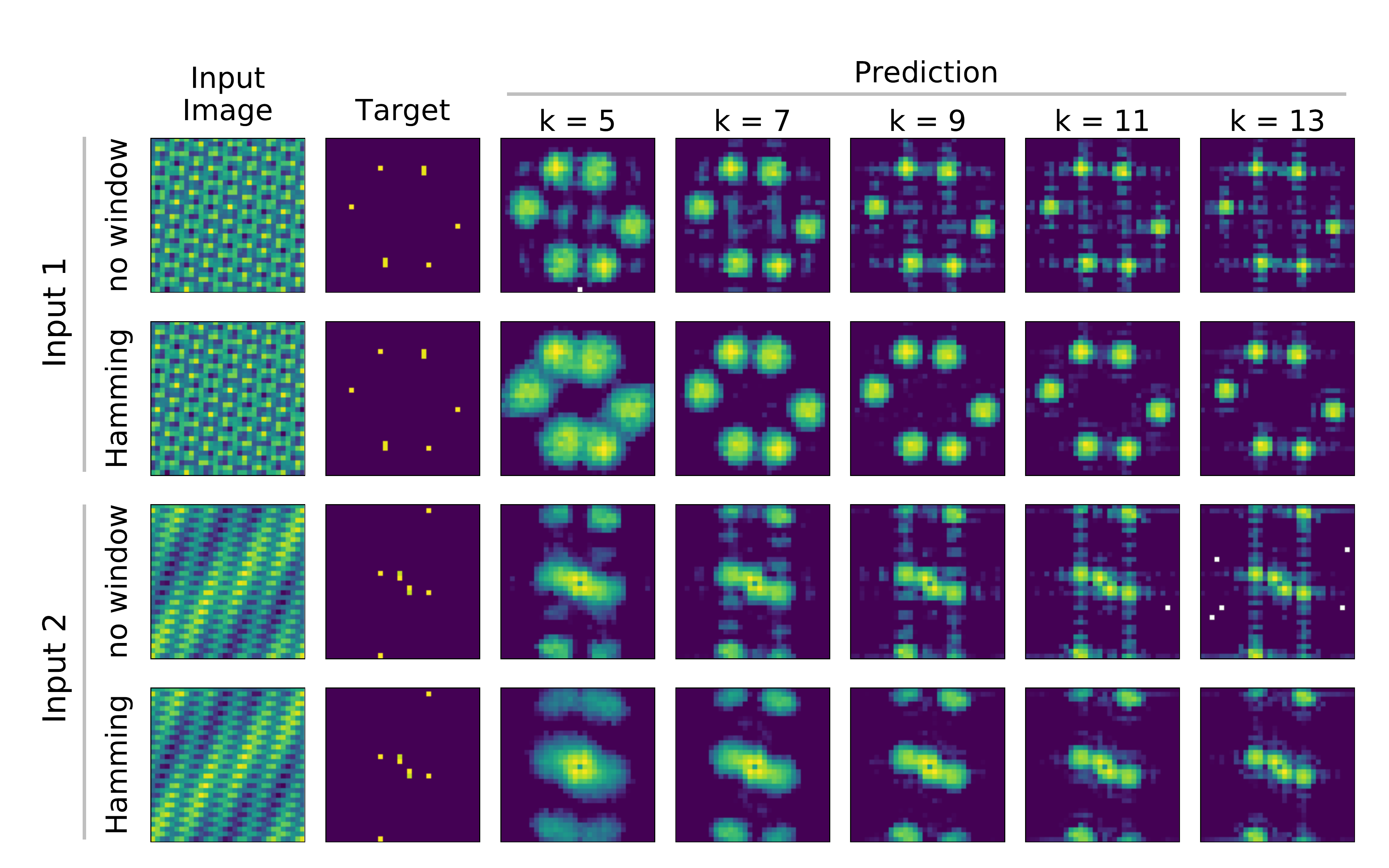}
    \end{center}
    \caption{Predictions of the baseline (no window) and Hamming networks with different kernel sizes $k$ for two example input images. The input images and the corresponding target FFTs are the same as those shown in Fig. 4. We find that the responses of different output channels, and hence the predictions of the models, become more localized with increasing $k$, as the central lobe of the filter responses become narrower. However, we find that baseline networks do not readily suppress leakage artifacts, which can appear far from target frequency bins, even when they corrupt the predictions. Hamming models, in contrast, are more successful in keeping the responses constrained around target frequencies.}
    \label{fig:supp_preds}
\end{figure}

The tapering performed by the Hamming window also works to attenuate spectral leakage in the frequency domain. This effect is demonstrated in the network predictions shown in Fig.~\ref{fig:supp_preds}. We find that as the possible frequency resolution obtainable by the filters increases in both networks with increasing kernel size $k$, the networks are able to produce more precise predictions, with maximum responses located around the target frequencies. However, we also find that baseline networks, while optimizing their kernels to function as precise bandpass filters, also learn filters which are prone to considerable spectral leakage. This observation adds support to our original hypothesis that conventional CNN filters are susceptible to leakage artifacts, as the baseline networks in the FFT regression task still suffer from them, even when they directly contribute to the loss function. In contrast, the Hamming models are more successful in suppressing leakage artifacts, which indicates that explicit regularization by a window function may be useful in convolutional architectures.

\section{Orthogonality in windowed convolutions}
Orthogonal convolutional neural networks~\cite{Wang2020} have recently been proposed as an approach which can alleviate, to some degree, problems in CNNs related to overparameterization~\cite{Han2016} or under utilization of model capacity~\cite{Cheung2019}. Orthogonal representations may reduce redundancies in the learned weights and lead to better performance through optimal capacity usage. In~\cite{Wang2020}, Wang~et~al. suggest that the correct, and effective, way to enforce orthogonality is not to impose orthogonality between individual kernels, but between the rows or columns of the doubly block-Toeplitz (DBT) matrix which defines the linear operation in a convolutional layer.

As explained in Section 4.4, spectral leakage may reduce the frequency-selectivity of the filters in a convolutional layer. Thus the network model may be more prone to learning redundant representations. We suggest that, by removing truncation artifacts, the Hamming window may enable the learning of more orthogonal filters, with less overlap in their frequency responses. Hence, CNNs with larger kernels with the Hamming window, may lead to both redundancy reduction, and performance increase, as well as to convolutional layers more similar to an orthogonal convolution.

In order to test this hypothesis, we construct the DBT matrix $\mathcal{K}$ using the weight tensor $K$ in each layer. Specifically, for a convolutional layer with $C$ input channels, $M$ output channels and kernel size $k$, the standard weight tensor $K \in \mathbb{R}^{M\times C \times k \times k}$ can be rearranged, such that the linear operator of the convolutional layer can be written as a matrix multiplication between a flattened input feature map $\mathbf{x} \in \mathbb{R}^{CHW}$ and the DBT matrix $\mathcal{K} \in \mathbb{R}^{(MH'W') \times (CHW)}$
\begin{equation}
    \large
    \mathbf{y}=\mathcal{K}\mathbf{x}
\end{equation}
where $\mathbf{y} \in \mathbb{R}^{(MH'W')}$ is the output feature map before nonlinearities or normalization. $H$ and $W$, and $H'$ and $W'$ denote the spatial dimensions of the input $\mathbf{x}$ and output $\mathbf{y}$, respectively.

For our models, we consider the row orthogonality condition for the DBT matrix
\begin{equation}\label{eq:sup_orth}
    \large
    \delta_{ij}=\langle \mathcal{K}_{i,\boldsymbol{\cdot}},\mathcal{K}_{j,\boldsymbol{\cdot}} \rangle=
    \begin{cases} 
        1, & \textrm{if } \; i=j \\
        0, & \textrm{else}
    \end{cases}
\end{equation}
which shows that, for an orthogonal convolution, each row vector $\mathcal{K}_{i,\boldsymbol{\cdot}}$, corresponding to the $i$-th row of the DBT matrix, must be orthogonal to every other row vector of the DBT matrix, so that every pairwise dot product $\delta_{ij}$ between the row vectors $\mathcal{K}_{i,\boldsymbol{\cdot}}$ and $\mathcal{K}_{j,\boldsymbol{\cdot}}$ is $0$ except when $i=j$.

Based on this, we would like to find how much the ResNet and VGG models trained on ImageNet deviate from orthogonal convolutions in each layer. Therefore, we first normalize each row of the DBT matrix $\mathcal{K}_{i,\boldsymbol{\cdot}}$ to a unit vector, so that $\langle \mathcal{K}_{i,\boldsymbol{\cdot}},\mathcal{K}_{i,\boldsymbol{\cdot}} \rangle=1$, then compute the dot product $\hat{\delta}_{ij}$ between every pair of rows $(i,j)$. We report the average deviation $D$ of the dot product from the orthogonality condition
\begin{equation}\label{eq:sup_dev}
    \large
    D=\frac{1}{N(N-1)}\, \sum_{i=1}^{N} \, \sum_{j \neq i} \, | \hat{\delta}_{ij} |
\end{equation}
for every convolutional layer, where $N=MH'W'$ is the number of rows of the DBT matrix. In order to illustrate the effects of training, we also compute $D$ for a randomly initialized version of each model, which we call the `chance' level.

\begin{figure}
    \hspace{-0.7cm}
    \begin{tabularx}{\linewidth}{cc}
        {\hspace{1cm}ResNet Models} & {\hspace{1cm}VGG Models} \\
        \includegraphics[trim=0 0 0 0, width=0.49\linewidth]{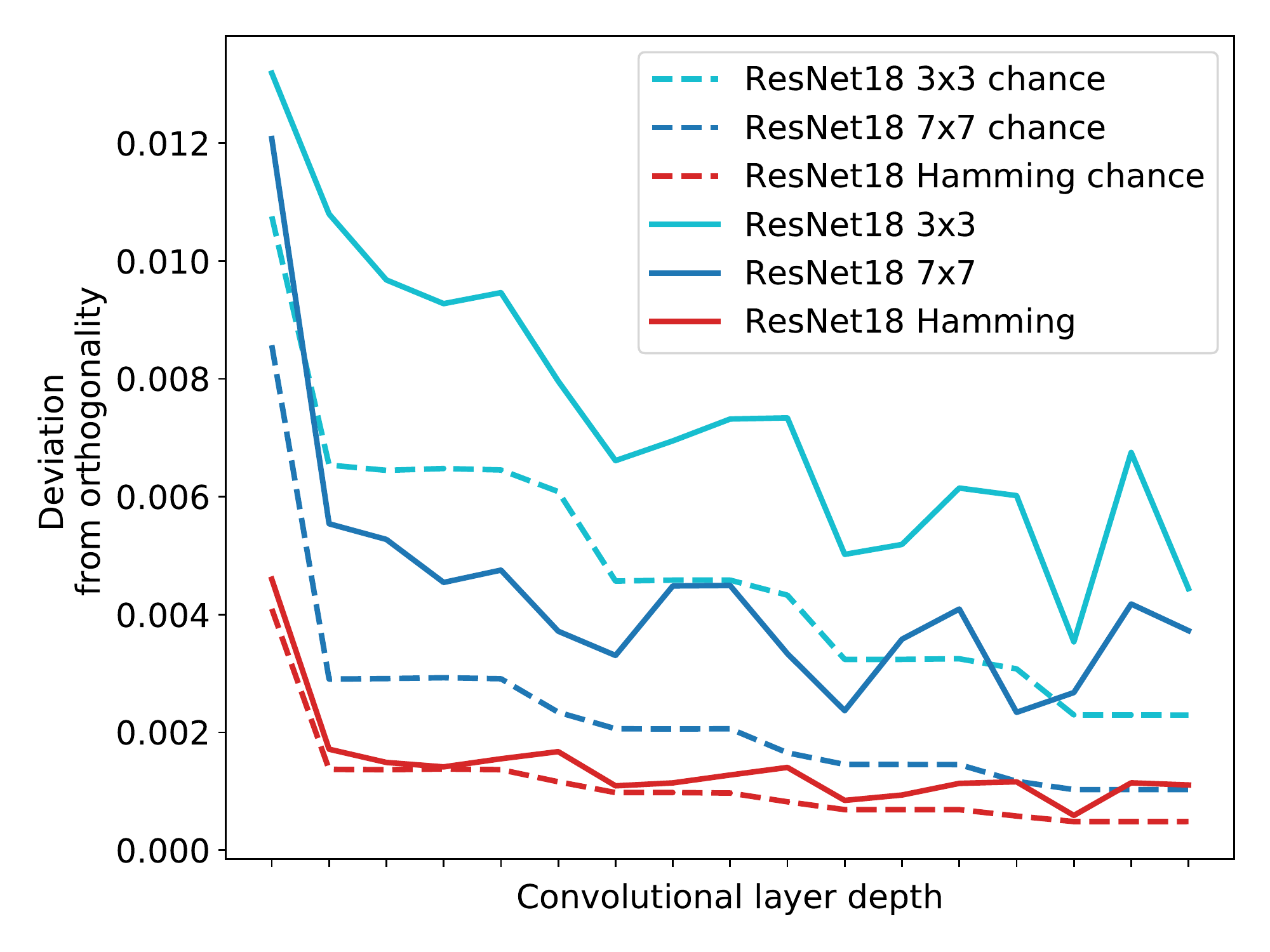} &
        \includegraphics[trim=0 0 0 0, width=0.49\linewidth]{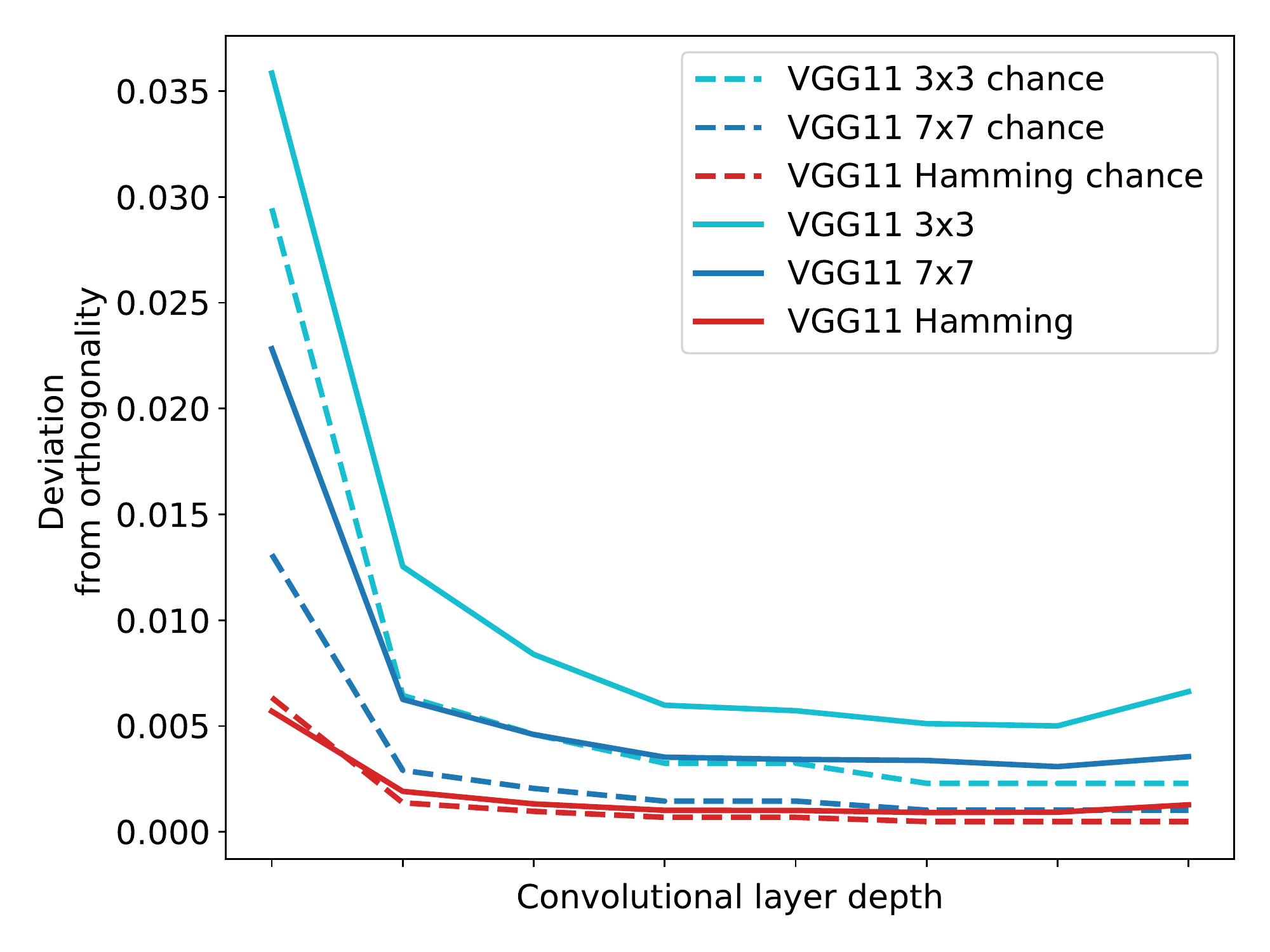} \\
        \includegraphics[trim=0 0 0 0, width=0.49\linewidth]{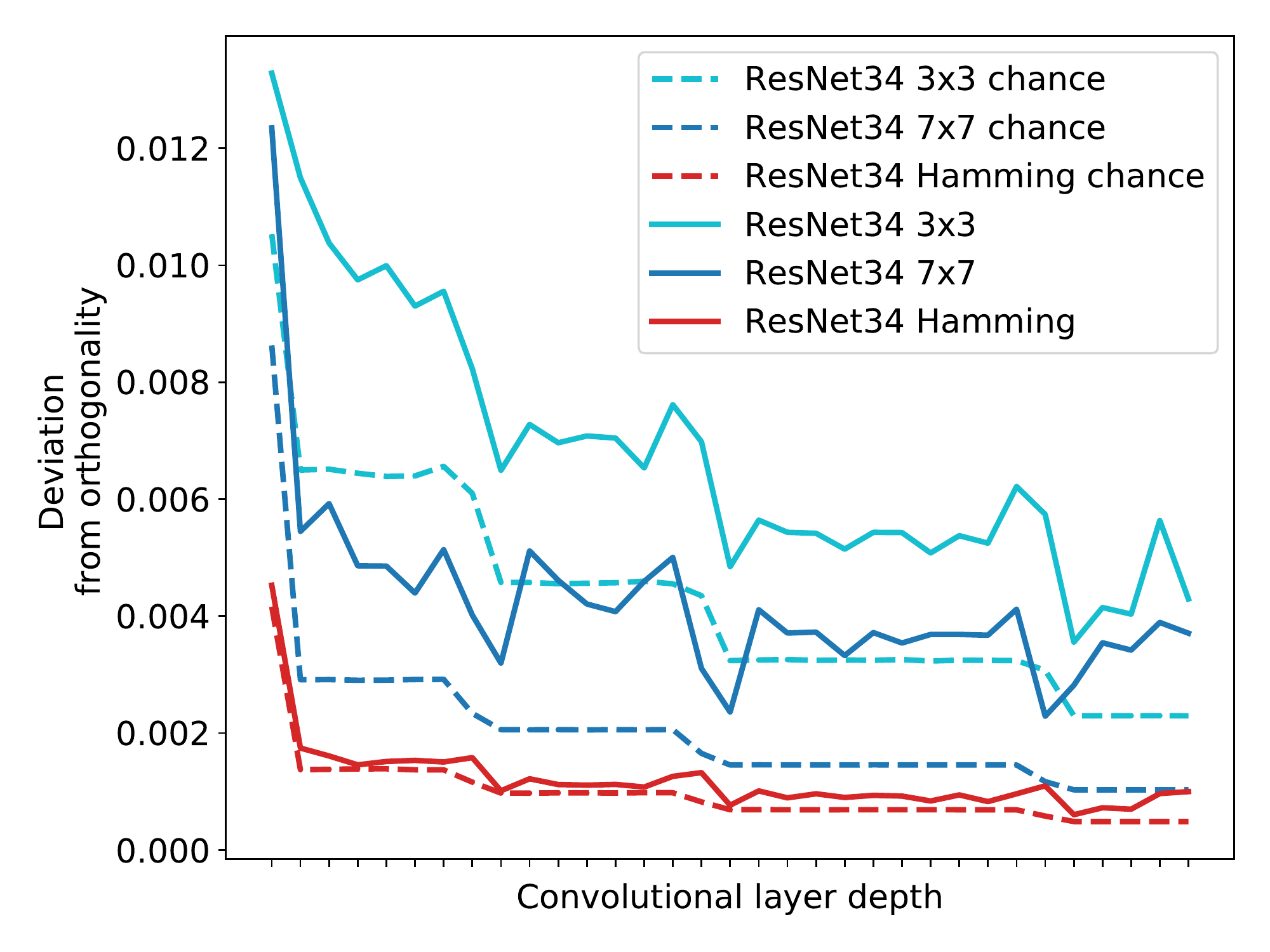} &
        \includegraphics[trim=0 0 0 0, width=0.49\linewidth]{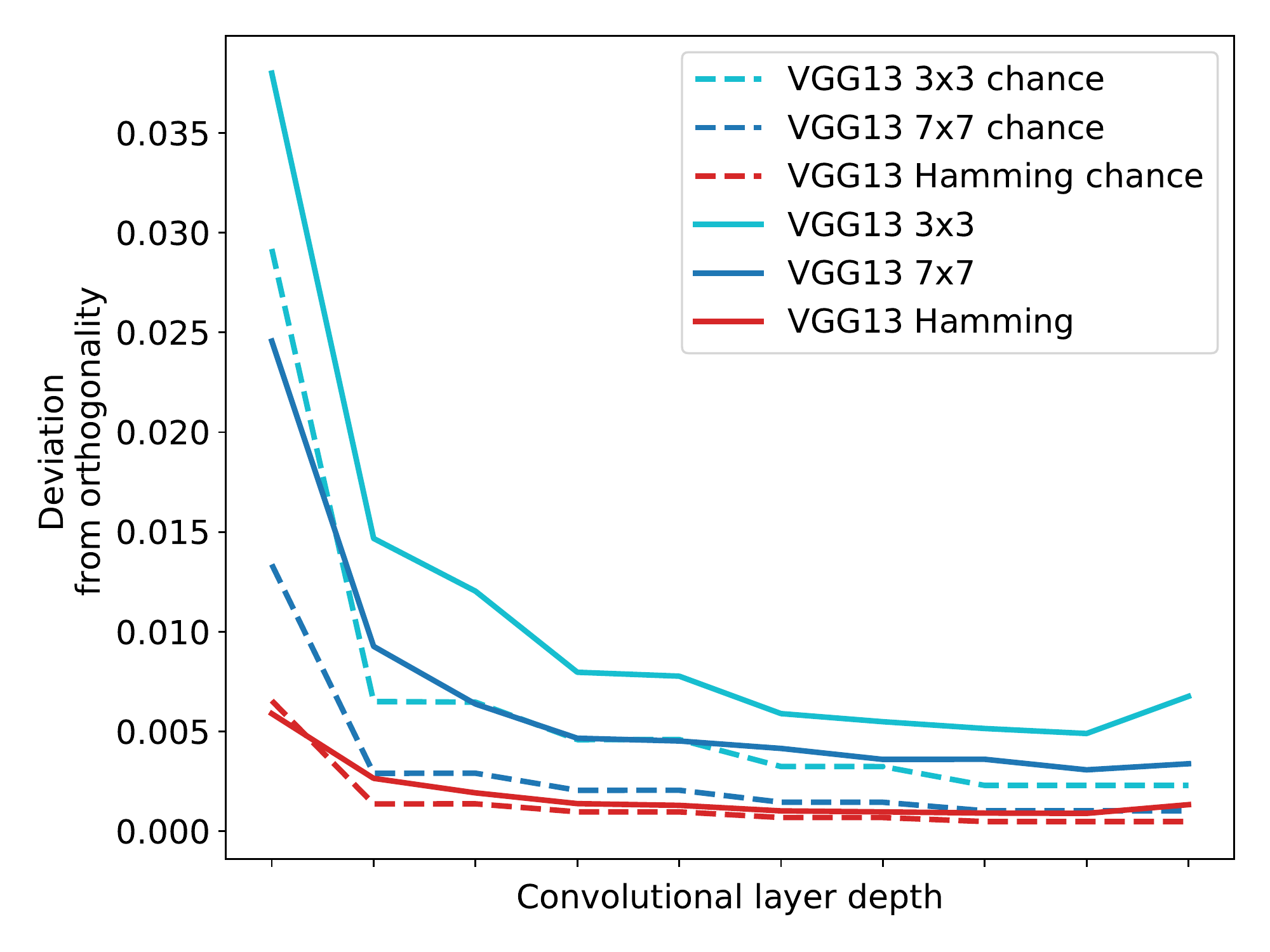} \\
        \includegraphics[trim=0 0 0 0, width=0.49\linewidth]{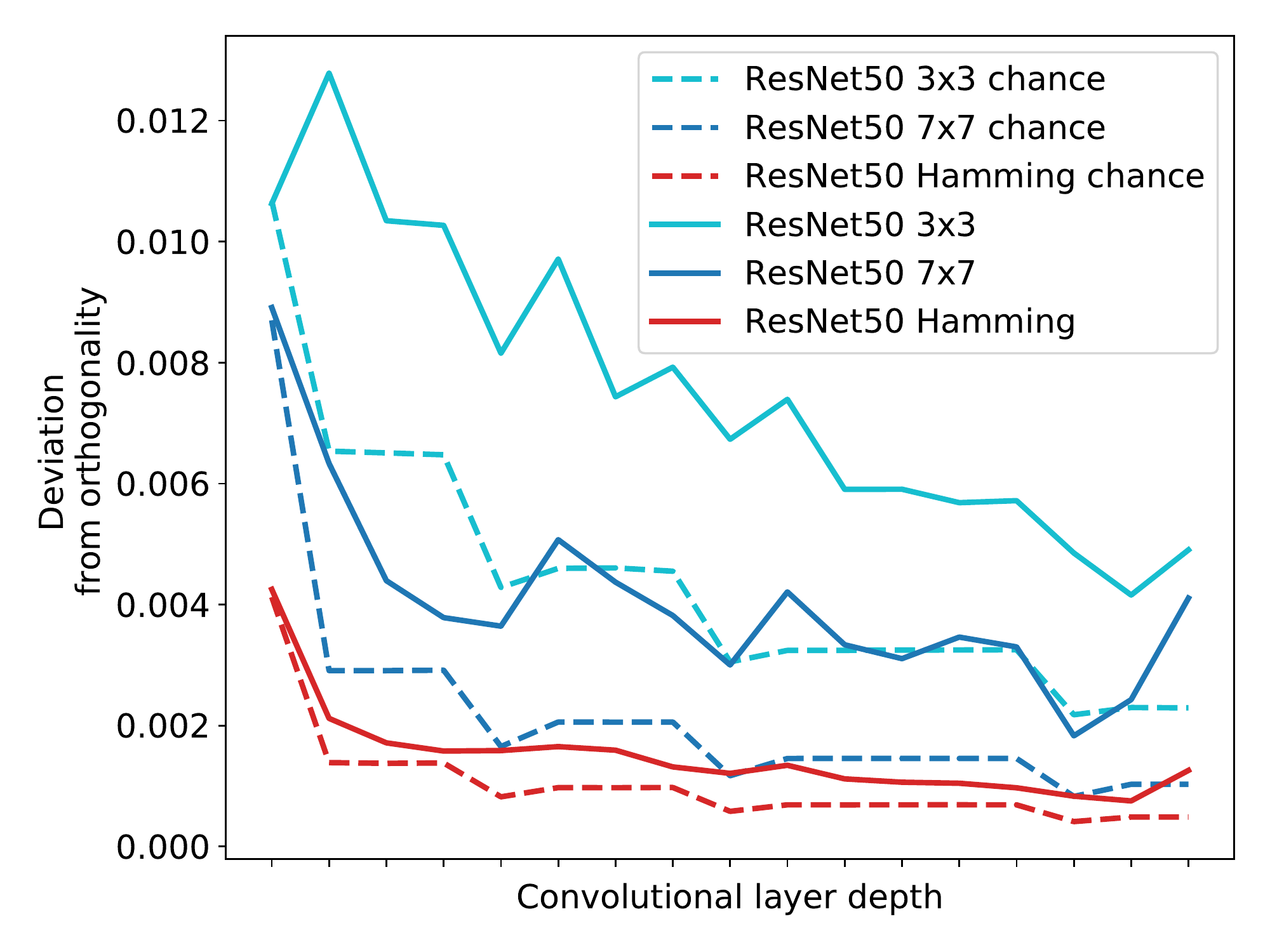} &
        \includegraphics[trim=0 0 0 0, width=0.49\linewidth]{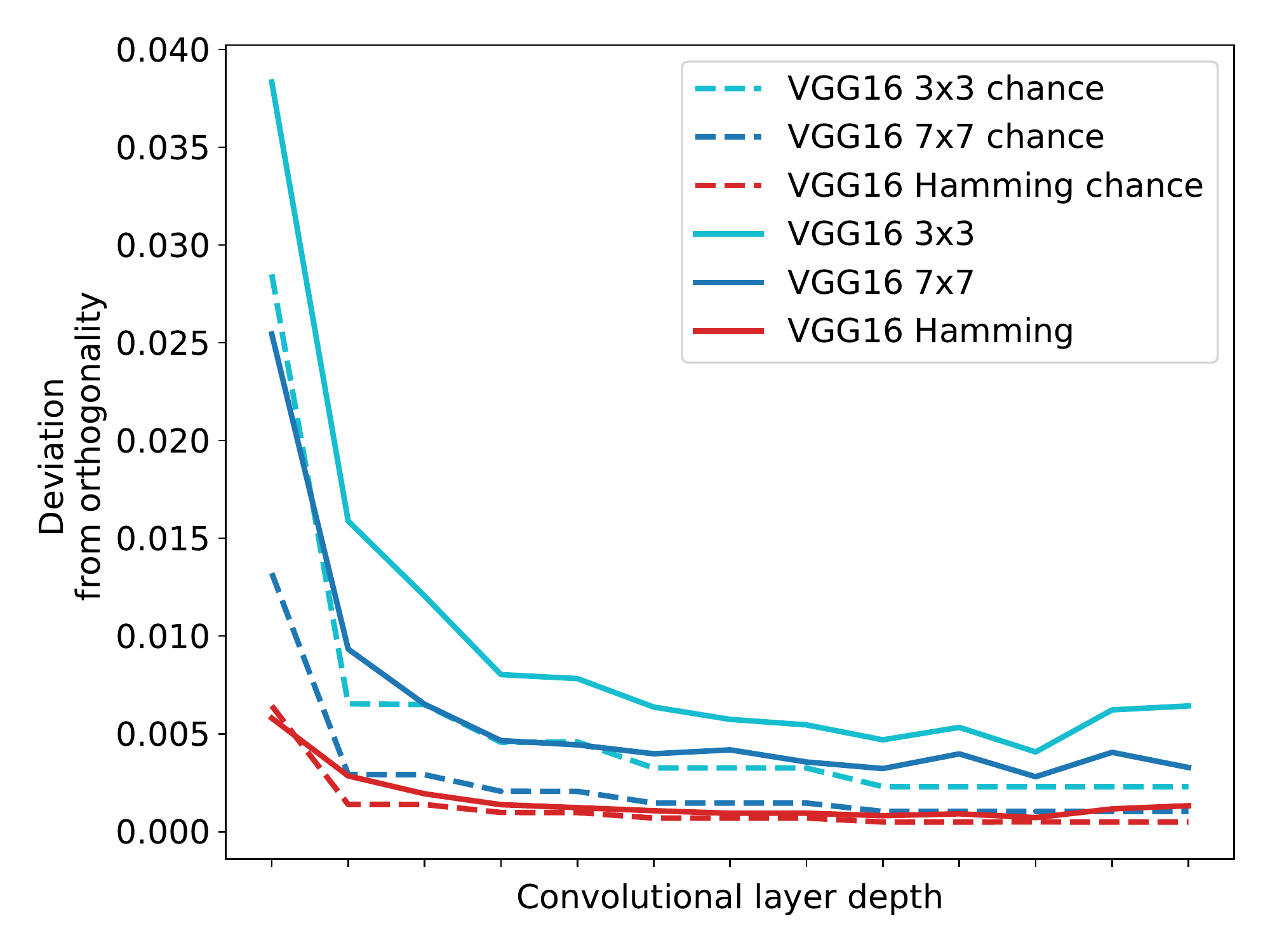} \\
    \end{tabularx}
    \caption{Deviation of the convolutional layers in different ResNet and VGG models from orthogonal convolutions~\cite{Wang2020}. The deviation $D$ is quantified as the average dot product between the row vectors of the doubly block-Toeplitz (DBT) matrix, as given in Eq.~\ref{eq:sup_orth}-\ref{eq:sup_dev}. A `chance' level is computed using a randomly initialized version of every model (dashed lines). We find that Hamming-windowed convolutions (red) deviate minimally from orthogonal convolutions as compared to baselines (cyan and blue). In addition, the increase in $D$ caused by training is the smallest for Hamming models.}
    \label{fig:supp_orthogonal}
\end{figure}

We find that the convolution operators deviate from orthogonality the least when using the Hamming window for all ResNet and VGG models~\ref{fig:supp_orthogonal}. For all baseline models, we find that training increases the deviation from orthogonality, and $D$ is on average higher than its corresponding chance level. However, we also find that this difference between the deviation $D$ of the trained model and its chance level is minimized for the Hamming models.

\vfill

\end{document}